\pdfoutput=1

\documentclass[11pt]{article}

\usepackage[final]{acl}

\usepackage{times}
\usepackage{latexsym}

\usepackage[T1]{fontenc}

\usepackage[utf8]{inputenc}

\usepackage{microtype}

\usepackage{inconsolata}

\usepackage{graphicx}

\usepackage{amsmath}
\usepackage{amssymb}
\usepackage{mathtools}
\usepackage{amsthm}

\usepackage[capitalize,noabbrev]{cleveref}

\usepackage{hyperref}
\theoremstyle{plain}
\newtheorem{theorem}{Theorem}[section]

\theoremstyle{definition}
\newtheorem{definition}[theorem]{Definition}

\theoremstyle{remark}

\usepackage{multirow}
\usepackage{multicol}
\usepackage{threeparttable}
\usepackage{makecell}
\usepackage{tabularx}

\usepackage{pifont}

\usepackage[export]{adjustbox}
\usepackage{subcaption}
\usepackage{graphicx}
\usepackage{wrapfig}

\usepackage{bbold}
\usepackage{enumitem}
\usepackage{booktabs}

\newcommand{\rev}[1]{\textcolor{black}{#1}}

\usepackage{bbm}

\usepackage{fancyhdr}
\fancypagestyle{firstpage}{
  \fancyhf{} %
  \fancyfoot[C]{{\small\itshape Accepted in Proceedings of the 2025 Conference on Empirical Methods in Natural Language Processing}} %
}

\usepackage{tcolorbox}

\usepackage{minitoc}

\usepackage[titles]{tocloft}

\title{Model-based Large Language Model Customization as Service}

\author{Zhaomin Wu$^{1*}$, Jizhou Guo$^{2\dagger*}$, Junyi Hou$^{1}$, {\bf Bingsheng He$^{1}$}, {\bf Lixin Fan$^{3}$}, {\bf Qiang Yang$^{4}$}\\
$^{1}$National University of Singapore, $^{2}$Zhiyuan Colleage, Shanghai Jiao Tong University\\
$^{3}$WeBank, $^{4}$The Hong Kong University of Science and Technology\\
\texttt{zhaomin@nus.edu.sg, sjtu18640985163@sjtu.edu.cn, hhh@u.nus.edu}\\
\texttt{dcsheb@nus.edu.sg, lixinfan@webank.com, qyang@cse.ust.hk}\\
}

\newcommand\blfootnote[1]{%
  \begingroup
  \renewcommand\thefootnote{}\footnote{#1}%
  \addtocounter{footnote}{-1}%
  \endgroup
}

\begin{document}
\doparttoc %
\faketableofcontents %

\maketitle
\thispagestyle{firstpage}

\begin{abstract}

Prominent Large Language Model (LLM) services from providers like OpenAI and Google excel at general tasks but often underperform on domain-specific applications. Current customization services for these LLMs typically require users to upload data for fine-tuning, posing significant privacy risks. While differentially private (DP) data synthesis presents a potential alternative, its application commonly results in low effectiveness due to the introduction of excessive noise on data for DP. To overcome this, we introduce \textit{Llamdex}, a novel framework that facilitates LLM customization as a service, where the client uploads pre-trained domain-specific \textit{models} rather than data. This client-uploaded model, optionally protected by DP with much lower noise, is inserted into the base LLM via connection modules. Significantly, these connecting modules are trained without requiring sensitive domain data, enabling clients to customize LLM services while preserving data privacy. Experiments demonstrate that Llamdex improves domain-specific accuracy by up to 26\% over state-of-the-art private data synthesis methods under identical privacy constraints and, by obviating the need for users to provide domain context within queries, maintains inference efficiency comparable to the original LLM service.

\end{abstract}

\blfootnote{$^{*}$Equal contribution. $^{\dagger}$Work done during an internship at National University of Singapore.}

\section{Introduction}\label{sec:introduction}

\begin{figure}[t!]
    \centering
    \includegraphics[width=\columnwidth]{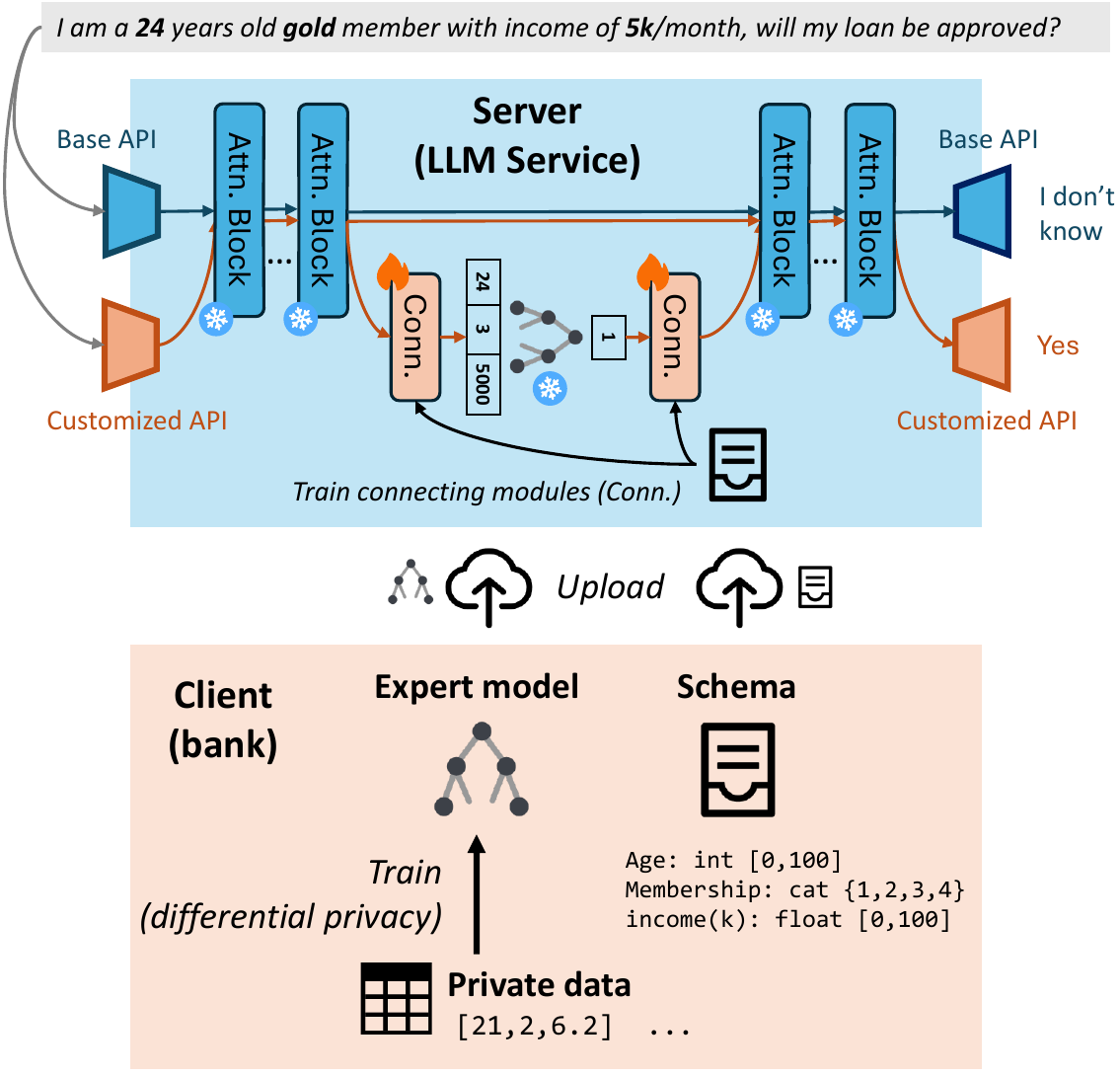}
    \caption{Overview of Llamdex customization pipeline}
    \label{fig:overview}
\end{figure}

While Large Language Model (LLM) services, such as Gemini~\cite{team2023gemini} and ChatGPT~\cite{openai2023gpt}, excel at general tasks, they often exhibit limitations in domain-specific applications due to insufficient access to relevant private data. Customizing these LLM services typically involves \textit{clients} uploading domain data to providers (\textit{servers}) like Google or OpenAI for fine-tuning~\cite{hu2021lora}, with the resulting customized LLMs often hosted on platforms such as Gemini's Gem and OpenAI's GPT Store. However, this data-upload requirement introduces significant privacy risks, deterring clients in sensitive sectors such as healthcare and finance.

Existing privacy-preserving approaches for LLM customization services often compromise on effectiveness to achieve privacy. A common method is differentially private (DP) data synthesis~\cite{hong2024dpopt,tian2022seqpate,duan2024flocks}. This technique involves clients uploading synthetic data, generated from their original private datasets under DP constraints, to the service provider. While this approach theoretically provides strong DP guarantees for the client's underlying domain data, the requisite noise injection for achieving such privacy significantly degrades synthetic data quality. This degradation, in turn, diminishes the LLM customization service's effectiveness, leading to customized LLM services that often exhibit notably reduced response accuracy on domain-specific tasks, a phenomenon termed ``disparate impact''~\cite{ganev2022robin}.

To improve this privacy-effectiveness trade-off in LLM customization services, we propose a novel framework named \textbf{L}arge \textbf{la}nguage \textbf{m}odel with \textbf{d}omain \textbf{ex}pert (\textit{Llamdex}). This framework enables clients to customize the LLM service by uploading a domain-specific \textit{model} instead of potentially sensitive \textit{data}, thereby offering superior privacy compared to data-sharing methods. An overview of the Llamdex service architecture, illustrated for financial inquiry applications, is presented in Figure~\ref{fig:overview}. This architecture involves the client providing a pre-trained, domain-specific model (termed the \textit{expert model}) to the server. The server then inserts this expert model into an intermediate layer of the base LLM via learnable connecting modules. Concurrently, a data schema detailing each column's name, type, and range is also supplied to the service, enabling the base LLM to correctly interpret the inputs and outputs of the client's expert model during service operation. 

This model-based customization approach for LLM services yields several advantages regarding effectiveness, privacy, and efficiency. First, \textbf{it enhances effectiveness by decoupling context understanding from task-solving processes}. The LLM can then focus on context understanding, its area of strength, while task-solving is delegated to the expert model, which can be any model (e.g., XGBoost~\cite{chen2016xgboost}) optimized for the specific task. This separation improves overall effectiveness, as LLMs often exhibit lower accuracy on tasks like arithmetic calculations~\cite{yuan2023well} or precise search~\cite{saparov2025transformers} where specialized models excel. Second, concerning privacy, the client's expert model can be trained using established DP techniques~\cite{abadi2016deep}. This approach \textbf{yields considerably lower noise than DP data synthesis methods}~\cite{tian2022seqpate} under an equivalent privacy budget~\cite{li2021large}. Finally, the Llamdex service architecture \textbf{maintains inference efficiency comparable to the base LLM}. Integrating the expert model avoids the need for users to embed extensive contextual information within each service prompt.

Concurrently, this design presents two significant challenges. The first is the misalignment between the domain expert's operational space (e.g., 1D feature vectors from tabular data) and the LLM's token embedding representations, necessitating complex design for trainable connecting modules. The second challenge is the absence of training data for these modules, since the client's original data is inaccessible due to privacy. Overcoming these obstacles is vital for successful model-based knowledge transfer.

To overcome these challenges, we design the architecture and training algorithm of Llamdex to facilitate effective model-based knowledge transfer. Specifically, to resolve the first challenge, we design \textit{Llamdex encoder} that maps the original tokens to feature vectors and a \textit{Llamdex decoder} that converts the expert's output into multiple token embeddings. These embeddings are then appended to the original sequence of token embeddings. To address the second challenge, we train the mapping modules using synthetic texts generated from randomly distributed data under the public schema. This allows the mapping modules to learn to extract the required feature values from texts without relying on the real data distribution. Our source code is available at a GitHub repository\footnote{\url{https://github.com/Xtra-Computing/Llamdex}}. The contributions of this paper are as follows:
\begin{itemize}[leftmargin=15pt]
    \item We propose a novel framework, Llamdex, that enhances LLM customization by effectively integrating domain-specific models into LLMs through trainable connecting modules.
    \item We design a training algorithm for connecting modules using only the public schema, without requiring access to domain data distributions in either original or perturbed form.
    \item Experiments on real-world datasets demonstrate Llamdex's superior customization effectiveness, achieving accuracy improvements of up to 14\% over LoRA fine-tuning and up to 26\% over PromptPATE (at an equivalent differential privacy level), while maintaining inference efficiency comparable to the base LLM.
\end{itemize}

\section{Background}\label{sec:background}

To provide essential background for comparing the privacy levels of different approaches, we briefly introduce the concept of differential privacy (DP).

\begin{definition}[Differential Privacy~\cite{dwork2006differential}]
    A randomized algorithm $\mathcal{M}: \mathcal{D} \rightarrow \mathcal{O}$ is said to be $(\varepsilon, \delta)$-differentially private if, for any two datasets $D, D' \in \mathcal{D}$ that differ by a single record, and for all output sets $S \subseteq \mathcal{O}$,
    \begin{equation}
        \Pr[\mathcal{M}(D) \in S] \leq e^{\varepsilon} \Pr[\mathcal{M}(D') \in S] + \delta,
    \end{equation}
where $\varepsilon$ is the privacy budget; smaller $\varepsilon$ implies stronger privacy guarantees. $\delta$ is the probability that the privacy guarantee is breached.
\end{definition}

DP guarantees are typically achieved by introducing Gaussian noise during computation. Applying this concept to train deep learning models, notably through the Differentially Private Stochastic Gradient Descent (DP-SGD) algorithm~\cite{abadi2016deep}, enables training private model.

\section{Related Work}\label{sec:related-work}

Privacy-preserving LLM customization primarily falls into two categories: data-based and API-based methods. Data-based approaches involve clients providing synthetic data (generated from private domain data under DP guarantees) to the LLM service for fine-tuning. Conversely, API-based methods enable LLM customization by having LLM query an external, domain-specific API at inference time.

\paragraph{Data-based Customization.}
Data-based approach is broadly applicable to a diverse range of clients, from individuals to large corporations, primarily because it typically does not necessitate significant client-side computational resources or require the client to maintain constant online connectivity with the LLM service. Typically, data-based approaches employ differentially private data synthesis. For instance, some methods, such as PATE~\cite{papernot2016semi} and SeqPATE~\cite{tian2022seqpate}, involve adding noise to aggregated predictions from an ensemble of teacher models (trained on private domain data) to create synthetic data. Other variants, including $d_\chi$-DP~\cite{feyisetan2020privacy}, selective-DP~\cite{shi2021selective}, Table Diffusion~\cite{truda2023generating}, PromptPATE~\cite{duan2024flocks}, and DP-OPT~\cite{hong2024dpopt}, aim to further enhance the quality of the generated synthetic data. This synthesized data is subsequently used to fine-tune the base LLM. A notable drawback of such approaches that train models from synthetic data (a form of \textit{input perturbation}) is the substantial degradation of model utility~\cite{jayaraman2018distributed} when compared to \textit{gradient perturbation} methods that add noise during the training. Llamdex, which customizes LLM with DP-trained model, aligns with gradient perturbation methods, offering higher effectiveness of customization.

\paragraph{API-based Customization.}
This method is generally more suitable for large organizations that possess sufficient resources to develop, host, and maintain these APIs---ensuring they are consistently online and responsive---and is typically not well-suited for individual users due to these requirements. The mechanisms behind such APIs vary; for example, \citet{yao2022react} require users to provide API documentation to the LLM at inference time, while \citet{schick2024toolformer} and \citet{qin2023toolllm} propose fine-tuning an auxiliary LLM on the client-side to process API calls. These approaches can demand significant computational resources on the client side, may incur notable communication latency, and require the client's API infrastructure to be perpetually online. In contrast, Llamdex is designed for more general applicability, catering to both individual clients and large companies. It obviates the need for constant client online presence; clients only need to train a relatively small domain-specific model (e.g., XGBoost~\cite{chen2016xgboost}) on their data and then upload this model to the server once for customization.

\section{Problem Definition}\label{sec:problem-statement}

Consider a collaboration between a \textit{server} and a \textit{client}. The server possesses an LLM $\mathcal{M}$, parameterized by $\theta_{\mathcal{M}}$, pre-trained on a large public text dataset $\mathbf{X}^s \in \mathbb{A}^{N_s \times l}$, where $\mathbb{A}$ denotes the vocabulary (token set) and $l$ is the sequence length. The client holds a private, domain-specific dataset $\mathbf{X}^c \in \mathbb{R}^{N_c \times m}$, where $m$ is the number of features. We assume the client's data is not contained within the server's training data ($\mathbf{X}^c \not\subseteq \mathbf{X}^s$) and exhibits a distinct distribution. Associated with $\mathbf{X}^c$ is a public \textit{schema} $\mathbf{S}^c$, detailing feature names, types, and ranges. An illustrative example of $\mathbf{S}^c$ from the \texttt{titanic} dataset is provided below:

\begin{flushleft}
\small %
\ttfamily %
\textbf{Example Features (from $\mathbf{S}^c$ for $\mathbf{X}^c$):}\\
Age: int [0,100]; \\
Pclass: category \{``1'',``2'',``3''\}\\
\textbf{Example Target (from $\mathbf{S}^c$ for $\mathbf{y}$):} \\
Survived: bool \{False, True\}\\
\end{flushleft}

The server aims to adapt its LLM $\mathcal{M}$ (assumed to be a standard decoder-only transformer~\citep{jiang2023mistral,touvron2023llama}) into a domain-customized model $\mathcal{M}^c$ for the client's domain $\mathbf{X}^c$ without directly accessing $\mathbf{X}^c$. Specifically, we focus on a two-stage, model-based customization framework. 
\textbf{(Client-side)} The client trains a domain-specific expert model $\mathcal{E}^c$, parameterized by $\theta_{\mathcal{E}^c}$, on their private structured data $\mathbf{X}^c$ (feature vectors) to predict a target variable $\mathbf{y}$. Optionally, for enhanced privacy, $\mathcal{E}^c$ can be trained via DP-SGD~\citep{abadi2016deep} to ensure $\theta_{\mathcal{E}^c}$ satisfies $(\varepsilon, \delta)$-differential privacy. The client shares only these parameters $\theta_{\mathcal{E}^c}$ with the server as a component for service enhancement. \textbf{(Server-side)} The server inserts the client-provided expert model $\mathcal{E}^c$ into its frozen LLM $\mathcal{M}$ by training lightweight \textit{connector} parameters $\theta_{\text{conn}}$. This integration aims to enable the resulting customized LLM service, $\mathcal{M}^c$, to accurately answer domain-specific natural language questions. Formally, given a dataset $\mathbf{Z}^c = \{(\mathbf{z}_i, y_i)\}$, where each $\mathbf{z}_i$ is a natural language question related to $\mathbf{X}^c$ and $y_i$ is the target answer, we aim to optimize $\theta_{\text{conn}}$ by minimizing a generative loss $\mathcal{L}_{\text{gen}}$ over $\mathbf{Z}^c$:
\begin{equation} \label{eq:domain_aware_training}
    \min_{\theta_{\text{conn}}} \mathbb{E}_{(\mathbf{z}_i, y_i) \sim \mathbf{Z}^c} \left[ \mathcal{L}_{\text{gen}}(\mathcal{M}^c(\theta_{\text{conn}}; \mathbf{z}_i, \theta_{\mathcal{M}}, \theta_{\mathcal{E}^c}), y_i) \right].
\end{equation}
Crucially, both the base LLM parameters $\theta_{\mathcal{M}}$ and the expert model parameters $\theta_{\mathcal{E}^c}$ remain frozen during this stage; only $\theta_{\text{conn}}$ is trained by the server.

Given that tabular data is prevalent in relational databases and readily translatable to/from text via schemas~\citep{jatana2012survey}, we focus primarily on tabular $\mathbf{X}^c$. Extensions of this customization service to other modalities (e.g., images, video) are discussed in Section~\ref{sec:future-work} as future work.

\paragraph{Threat Model.}
We focus on the privacy of client data $\mathbf{X}^c$ against a semi-honest server that, while adhering to the protocol, may attempt to infer $\mathbf{X}^c$. The server accesses the LLM parameters $\theta_{\mathcal{M}}$, the client's schema $\mathbf{S}^c$, and the expert model parameters $\theta_{\mathcal{E}^c}$. The primary privacy risk involves the server inferring $\mathbf{X}^c$. If schema $\mathbf{S}^c$ is private, masking techniques~\citep{ranganathan2023tool} can offer protection, an aspect orthogonal to this study.

\section{Approach}\label{sec:approach}

This section details the Llamdex design for LLM service customization. Our approach uses two types of client-provided information: client’s data distributions (via a client-trained expert model $\theta_{\mathcal{E}^c}$, optionally trained with DP) and client's data schema $\mathbf{S}^c$. The server uses $\mathbf{S}^c$ to train the connecting modules that bridge the frozen base LLM and the expert model, enabling the customized service. Section~\ref{subsec:model_architecture} describes this model architecture, focusing on the Llamdex encoder/decoder design. Section~\ref{subsec:training_inference} details the schema-guided training and the service's inference procedure. Finally, Section~\ref{subsec:iterative_reasoning} introduces service extensions for generating explanations and enabling iterative reasoning.

\begin{figure}[htpb]
    \centering
    \includegraphics[width=\linewidth]{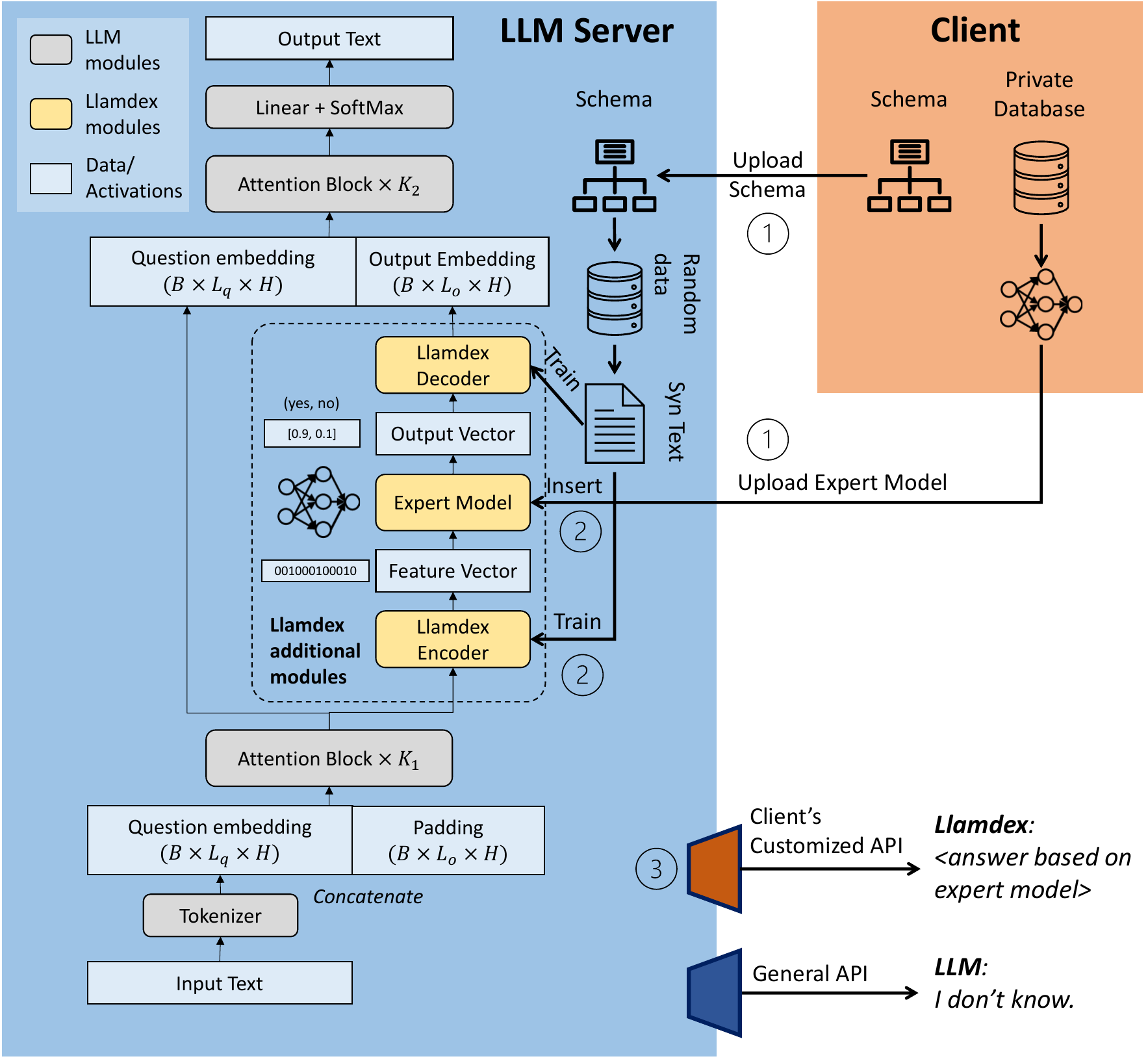}
    \caption{Llamdex structure and customization pipeline}
    \label{fig:llamdex-full}
\end{figure}

\subsection{Model Architecture}\label{subsec:model_architecture}

Llamdex facilitates LLM service customization by inserting a module between two attention blocks (assumed to be the $k$-th block) of the frozen base LLM $\mathcal{M}$, as depicted in Figure~\ref{fig:llamdex-full}. This module processes LLM embeddings $\mathbf{h}^k_i$ derived from the input question $\mathbf{z}_i$ and generates domain-informed \textit{output embeddings} $\mathbf{O}^{\text{emb}}_i$, which are then passed to subsequent LLM layers. This module comprises three core components: an expert model $\mathcal{E}^c$, a \textit{Llamdex encoder}, and a \textit{Llamdex decoder}. To maintain sequence length consistency required by some LLM architectures (e.g., those using RoPE~\citep{su2024roformer}), Gaussian noise paddings is appended to $\mathbf{h}^k_i$ before reaching the expert model, which are subsequently replaced by $\mathbf{O}^{\text{emb}}_i$.

\paragraph{Expert Model.}
The expert model $\mathcal{E}^c$, parameterized by the client-provided $\theta_{\mathcal{E}^c}$, encapsulates information about the client's domain-specific data distribution. It accepts a feature vector $\mathbf{x}^c_i$ and outputs a prediction $\hat{y}_i = \mathcal{E}^c(\theta_{\mathcal{E}^c}; \mathbf{x}^c_i)$. As $\theta_{\mathcal{E}^c}$ is frozen on the server-side, $\mathcal{E}^c$ can be any suitable model (e.g., multi-layer perceptron (MLP) or XGBoost), optionally trained with DP guarantees by the client.

\begin{figure}[htpb]
    \centering
    \begin{subfigure}[b]{0.45\linewidth}
    \centering
    \includegraphics[width=0.95\linewidth]{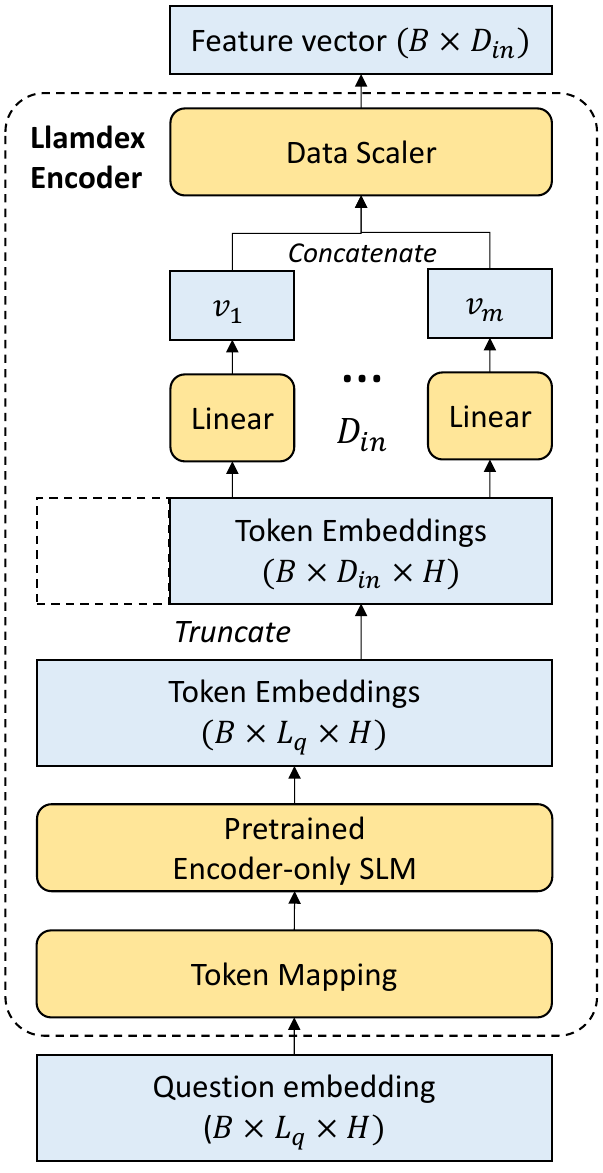}
    \caption{Llamdex Encoder}
    \label{fig:llamdex-encoder}
    \end{subfigure}
    \begin{subfigure}[b]{0.45\linewidth}
    \centering
    \makebox[\linewidth][r]{%
    \includegraphics[width=0.65\linewidth]{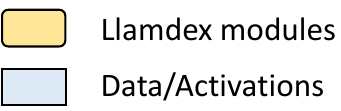}
    }
    \vspace{27pt}\par %
    \includegraphics[width=0.95\linewidth]{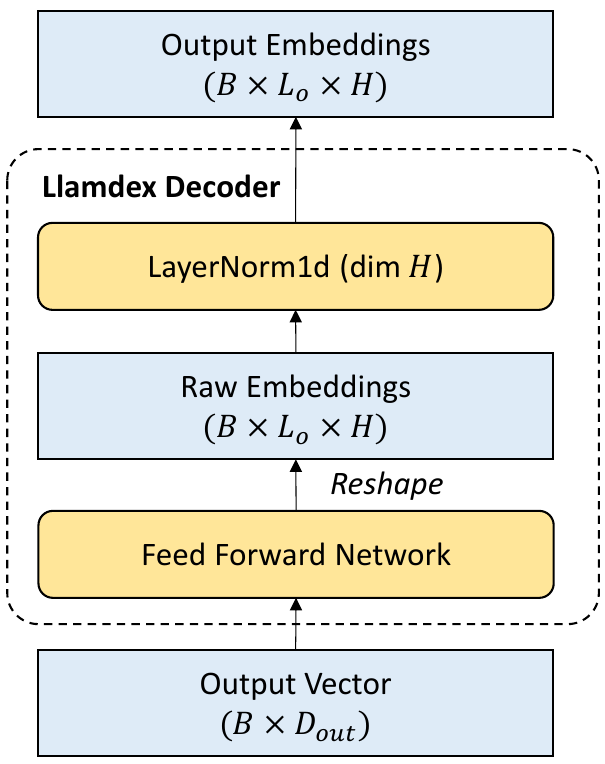}
    \caption{Llamdex Decoder}
    \label{fig:llamdex-decoder}
    \end{subfigure}
    \caption{Llamdex encoder and decoder architecture}
    \label{fig:llamdex-encoder-decoder}
\end{figure}

\paragraph{Llamdex Encoder.}
The Llamdex encoder translates intermediate question embeddings from the LLM $\mathcal{M}$ into the structured feature vector $\mathbf{x}^c_i$ required by the expert model $\mathcal{E}^c$ (see Figure~\ref{fig:llamdex-encoder-decoder}). The primary challenge stems from the auto-regressive nature of decoder-only LLMs; these models are optimized to predict the next token one at a time based on prior context, which complicates the direct, single-step extraction of a complete, structured feature set from their hidden states. To address this inherent limitation, we employ a pre-trained encoder-only small language model (SLM), e.g. RoBERTa~\cite{liu2019roberta}, for feature extraction.

A significant hurdle when using an auxiliary SLM is the potential misalignment between the tokenizers of the main LLM ($\mathcal{M}$) and the SLM. To overcome this, we introduce a \textit{token mapping} module, inspired by the logit lens~\cite{nostalgebraist2020logitlens} which was originally proposed for token interpretation. Specifically, given intermediate LLM question embeddings $\mathbf{h}^k_i$ at $k$-th attention block, the token mapping module computes logits $\mathbf{\Lambda}_i$ using the $\text{RMSNorm}$ layer and LLM head $\mathcal{H}_{\text{LLM}}$ of $\mathcal{M}$:
\begin{equation}
    \mathbf{\Lambda}_i = \mathcal{H}_{\text{LLM}}(\text{RMSNorm}_{\text{LLM}}(\mathbf{h}^k_i)).
\end{equation}
For each token, the most probable token ID $\boldsymbol{\tau}^{\text{LLM}}_i$ according to the LLM are extracted:
\begin{equation}
    \boldsymbol{\tau}^{\text{LLM}}_i = \arg\max(\mathbf{\Lambda}_i).
\end{equation}
These LLM token IDs are then decoded into text tokens using the LLM's tokenizer $\mathcal{T}_{\text{LLM}}$, formally, $\mathbf{t}^{\text{LLM}}_i = \mathcal{T}_{\text{LLM}}.\text{Decode}(\boldsymbol{\tau}^{\text{LLM}}_i)$. Finally, these text tokens are re-encoded using the SLM's tokenizer $\mathcal{T}_{\text{SLM}}$ to obtain token IDs compatible with the SLM, i.e., $\boldsymbol{\tau}^{\text{SLM}}_i = \mathcal{T}_{\text{SLM}}.\text{Encode}(\mathbf{t}^{\text{LLM}}_i)$. These aligned SLM token IDs $\boldsymbol{\tau}^{\text{SLM}}_i$ are embedded into $\mathbf{H}^{\text{SLM}}_i$ with the SLM. Then, a truncated sequence $\tilde{\mathbf{H}}^{\text{SLM}}_i$ is obtained by keeping the last $D_{\mathrm{in}}$ tokens embeddings; these token embeddings are projected into $D_{\mathrm{in}}$ feature values of $\mathbf{x}^c_i\in\mathbb{R}^{D_{\mathrm{in}}}$ via linear layers, followed by min-max scaling activations based on the ranges specified in $\mathbf{S}^c$, ensuring features adhere to valid ranges consistent with $\mathcal{E}^c$'s training.

\paragraph{Llamdex Decoder.}
The Llamdex decoder maps the expert model's prediction $\hat{y}_i$ back into the LLM's embedding space, generating the output embeddings $\mathbf{O}^{\text{emb}}_i$. The mapping is achieved using a simple feed-forward network (FFN) with SwiGLU activation~\citep{shazeer2020glu}. A key design challenge is the scale mismatch between these generated $\mathbf{O}^{\text{emb}}_i$ and the LLM's internal hidden states (e.g., $\mathbf{h}^k_i$ from preceding layers). A large discrepancy can destabilize subsequent computations, particularly skip-connections and attention mechanisms. To mitigate this, we apply an independent Layer Normalization (LayerNorm) specifically to the decoder's output before it is appended to the LLM's representations. This LayerNorm allows the model to adaptively learn the appropriate scale for $\mathbf{O}^{\text{emb}}_i$ that aligns with $\mathbf{h}^k_i$. The output embeddings $\mathbf{O}^{\text{emb}}_i$ are then appended to the question embeddings $\mathbf{h}^k_i$ and passed to the subsequent LLM layers.

\subsection{Training and Inference}\label{subsec:training_inference}

Llamdex service customization follows a two-stage protocol separating client and server responsibilities. \textbf{Client-side:} The client trains an expert model $\mathcal{E}^c$ on their private data $\mathbf{X}^c$, optionally protected by DP, and shares $\mathcal{E}^c$ and the public schema $\mathbf{S}^c$ with the server. \textbf{Server-side:} The server trains the connector parameters $\theta_{\text{conn}}$ (which comprise the Llamdex encoder and Llamdex decoder) using only the client's schema $\mathbf{S}^c$ and the received expert model parameters $\theta_{\mathcal{E}^c}$. Both the base LLM $\mathcal{M}$ and the client's expert model $\mathcal{E}^c$ remain frozen during this server-side training phase.

\paragraph{Training Llamdex Encoder.}
Since the server does not access the client's actual private data $\mathbf{X}^c$, the Llamdex encoder is trained using synthetically generated data. Based solely on the public schema $\mathbf{S}^c$ provided by the client, the server creates synthetic tabular feature vectors $\bar{\mathbf{x}}^c_i$ (e.g., via \texttt{sklearn}). Crucially, $\bar{\mathbf{x}}^c_i$ adheres to the schema's structure (types, ranges) but is generated from a \textbf{completely random distribution}, independent of the client's true data distribution underlying $\mathbf{X}^c$. For each $\bar{\mathbf{x}}^c_i$, a corresponding natural language questions $\bar{\mathbf{z}}_i$ is generated using an auxiliary LLM (e.g., Mistral-7B~\citep{jiang2023mistral}). The Llamdex encoder is then trained to map $\bar{\mathbf{z}}_i$ to $\bar{\mathbf{x}}^c_i$ with these synthetic data. This optimization uses a Mean Squared Error (MSE) loss, teaching the encoder the mapping from textual descriptions (e.g., \texttt{``He is ten years old''}) to structured features (e.g., \texttt{``10''}). Our experiments show that the Llamdex encoder's learned mapping $\bar{\mathbf{z}}_i\rightarrow\bar{\mathbf{x}}^c_i$ generalizes to $\mathbf{z}_i\rightarrow\mathbf{x}^c_i$ without knowing the distribution of $\mathbf{z}_i$ or $\mathbf{x}^c_i$.

\paragraph{Training Llamdex Decoder.}
Similarly, the Llamdex decoder is trained by the server using synthetically generated data. It learns to map the prediction $\hat{y}_i$ of $\mathcal{E}^c$ to LLM's embedding embeddings that finally produces the correct textual answer $a_i$ (e.g., \texttt{``Yes''}, \texttt{``No''}). To achieve this without accessing real label distribution, the server generate synthetic target labels $\bar{y}_i$ and corresponding textual answers $\bar{a}_i$, consistent with the target definition in the client's schema $\mathbf{S}^c$ (e.g., type and range). The Llamdex decoder is trained using a cross-entropy loss between its predicted logits and the ground truth logits of these synthetic textual answers $\bar{a}_i$. Our experiments find the $\bar{y}_i\rightarrow\bar{a}_i$ mapping learned by the decoder generalizes well to $\hat{y}_i\rightarrow a_i$ without being aware of the distribution of $y_i$ or $a_i$. 

\paragraph{Inference for Customized Service.}
The Llamdex system provides an end-to-end customized LLM inference service. Upon receiving a user's natural language question $\mathbf{z}_i$, the base LLM $\mathcal{M}$ processes it up to the $k$-th attention block, which is the insertion point for the integrated module. Here, the trained Llamdex encoder extracts the relevant feature vector $\mathbf{x}^c_i$ from the LLM's intermediate hidden states $\mathbf{h}^k_i$. This vector $\mathbf{x}^c_i$ is then fed to the client's expert model $\mathcal{E}^c$, which produces a domain-specific prediction $\hat{y}_i$. Subsequently, the trained Llamdex decoder transforms this prediction $\hat{y}_i$ into output embeddings $\mathbf{O}^{\text{emb}}_i$. These embeddings are appended to $\mathbf{h}^k_i$ and propagated to the LLM's subsequent layers to generate the final textual response. An illustrative use case is provided in Table~\ref{tab:text-data-example} (Appendix~\ref{sec:exp-detail}). Notably, Llamdex does not require either contextual information or the schema $\mathbf{S}^c$ at inference time, resulting in better efficiency than API-based services~\citep{yao2022react}.

\subsection{Iterative Reasoning Mechanism}\label{subsec:iterative_reasoning}

While the primary objective of Llamdex is to enable effective LLM customization, allowing the customized LLM to generate accurate answers based on the client's model, enhancing these answers with explanations or reasoning offers an additional service utility. To facilitate this, Llamdex incorporates a straightforward \textit{iterative feedback} mechanism, providing users with a method to obtain basic explanations for the service's outputs.

This mechanism allows users to request clarification by augmenting their initial query. Specifically, the original answer generated by the Llamdex service is concatenated with a predefined prefix prompt (e.g., \texttt{"The expert's answer is"}) and then appended to the original question $\mathbf{z}_i$. Users have the option to further refine this augmented query with custom prompts. The resulting revised input is subsequently resubmitted to the base LLM $\mathcal{M}$ component of the service. This base LLM processes the augmented context to produce a new response, aiming to deliver the requested reasoning or elaboration beyond the answer. An illustrative example of this iterative feedback mechanism is presented in Table~\ref{tab:reason} in Appendix~\ref{sec:add-exp}.

\section{Experiment}\label{sec:experiment}

This section presents our experimental evaluation. We begin by detailing the experimental settings (Section~\ref{sec:exp-setting}), followed by an assessment of the proposed method's accuracy (Section~\ref{sec:exp-accuracy}), inference efficiency (Section~\ref{sec:exp-efficiency}), and privacy guarantees (Section~\ref{sec:exp-privacy}). Supplementary analyses, including a comparison of training efficiency (Appendix~\ref{sec:train-efficiency}), additional experimental results such as those for iterative reasoning (Appendix~\ref{sec:add-exp}), comprehensive ablation studies (Appendix~\ref{sec:ablation}), and detailed hyperparameter configurations of baselines (Appendix~\ref{sec:baseline-hyper}), are provided in the appendices.

\subsection{Experiment Setting}\label{sec:exp-setting}

This subsection outlines experimental setups, including datasets, model configuration, evaluation methods, baselines, and the environment.

\paragraph{Dataset.} Our experiments incorporate four public real-world datasets: \texttt{titanic}~\cite{titanic}, \texttt{wine}~\cite{wine}, \texttt{bank}~\cite{bank}, and \texttt{nursery}~\cite{nursery}. For evaluation, all datasets are in tabular format with well-defined schema and meaningful column names. We split each dataset into training and test sets by 8:2. The details of each dataset are included in Appendix~\ref{sec:exp-detail}.

\paragraph{Expert Model.}
The expert models $\mathcal{E}_c$ are trained directly on the respective tabular datasets, utilizing MLPs (default) and XGBoost~\cite{chen2016xgboost}. MLPs, implemented in \texttt{PyTorch}~\cite{paszke2019pytorch}, feature two hidden layers (400/200 neurons, ReLU activation~\cite{nair2010rectified}), trained with AdamW~\cite{loshchilov2017adamw} (learning rate $10^{-4}$, batch 64, max 30 epochs or convergence). For DP expert model training, we adopt the DP-SGD implementation in \texttt{Opacus}~\cite{yousefpour2021opacus} library, which clips gradients by norm 1 and injects noise, with varying $\varepsilon$ and $\delta=1/N_c$, where $N_c$ is the number of samples. XGBoost models are configured with the following core hyperparameters: a learning rate of 0.1, \texttt{max\_depth} set to 50, \texttt{subsample} of 0.8, \texttt{colsample\_bytree} of 0.8, and \texttt{n\_estimators} set to 50.

\paragraph{LLM and SLM.} In our experiments, we employ the pretrained \texttt{Mistral-7B}~\cite{jiang2023mistral} as the base LLM, and the pretrained \texttt{Roberta-large}~\cite{meta2019roberta}, with 0.355 billion parameters, as the SLM within the Llamdex encoder. In {Llamdex} training, the LLM remains frozen while the SLM undergoes full-parameter fine-tuning. The Llamdex encoder and decoder are trained with a batch size of 128 and a learning rate of $5 \times 10^{-5}$, using the AdamW optimizer~\cite{loshchilov2017adamw}. The learning rate is adjusted by a cosine scheduler with 500 steps of warmup. The Llamdex encoder and decoder are trained for 30 and 10 epochs, respectively.

\paragraph{Evaluation.}
The effectiveness of customization is evaluated by the accuracy of answering domain-specific questions derived from client's tabular datasets. For question generation, each dataset row with $m$ columns, with 10\% of its values randomly masked to simulate missing data, is converted to text (format: $\#c_1: v_1, \dots, \#c_m: v_m$, where $c_i$ is column name, $v_i$ is value). This textual representation is then fed to \texttt{Mistral-7B} with a system prompt to generate a corresponding question. During accuracy evaluation, the LLM provides a single-word answer: ``Yes''/``No'' for binary classification or an uppercase letter (e.g., ``A''-``Z'') for multiclass tasks. The predicted class with the highest probability is compared against the ground truth to compute accuracy. we report the mean accuracy of five independent runs.

\paragraph{Baselines.}
To evaluate the accuracy and efficiency of Llamdex, we compare it against baselines without DP guarantees, including:
\begin{itemize}[leftmargin=15pt]
    \item \textbf{Original LLM}: Unmodified base LLM.
    \item \textbf{Real Data LoRA}~\cite{hu2021lora}: LLM parameter-efficiently fine-tuned (LoRA) on questions/labels derived from real domain data.
    \item \textbf{Expert API}: A simplified API-based approach~\cite{schick2024toolformer,qin2023toolllm} where the LLM is prompted to extract feature values from questions; values are regex-extracted and fed to a domain expert model that predicts the final answer.
\end{itemize}
To evaluate Llamdex's accuracy under privacy guarantees, we compare it with baselines employing DP data synthesis. The synthetic data generated by these methods is subsequently used for LoRA fine-tuning of the LLM. These baselines include:
\begin{itemize}[leftmargin=15pt]
    \item \textbf{PATE-GAN}~\cite{jordon2018pate}: Traditional DP tabular data synthesis.
    \item \textbf{SeqPATE}~\cite{tian2022seqpate}: DP data synthesis method using knowledge distillation.
    \item \textbf{PromptPATE}~\cite{duan2024flocks}: State-of-the-art DP data synthesis using private prompts.
    \item \textbf{Table Diffusion}~\cite{truda2023generating}: Diffusion-based DP tabular data generation.
    \item \textbf{DP-OPT}~\cite{hong2024dpopt}: DP prompt generation using an ensemble of 205 LLMs.
\end{itemize}

\paragraph{Environment.} Evaluations utilize a system with 4x NVIDIA H100 GPUs (80GB each) and an AMD EPYC 9654 96-Core processor with 1.11TB of CPU memory (large CPU memory is unnecessary).

\begin{table*}[t!]
    \centering
    \small
    \vspace{-5pt}
    \caption{Inference Time (minutes) and Peak Inference Memory Cost (GB)}
    \label{tab:inference}
    \vskip 0.15in
    \begin{tabular}{c cccc c cccc c}
        \toprule
        \multirow{2}{*}{\textbf{Method}} & \multicolumn{5}{c}{\textbf{Inference Time (min)}} & \multicolumn{5}{c}{\textbf{Inference Memory (GB)}} \\
        \cmidrule(lr){2-6} \cmidrule(lr){7-11}
        & \textbf{titanic} & \textbf{wine} & \textbf{bank} & \textbf{nursery} & \makecell{\textbf{Relative}\textsuperscript{1}} & \textbf{titanic} & \textbf{wine} & \textbf{bank} & \textbf{nursery} & \makecell{\textbf{Relative}\textsuperscript{1}} \\
        \midrule
        Original LLM & 0.03 & 0.40 & 2.58 & 0.72 & $\times$0.49 & 18.26 & 19.76 & 20.29 & 20.45 & $\times$0.96 \\
        \midrule
        \makecell{Syn. Data LoRA\textsuperscript{2}} & 0.03 & 0.42 & 2.75 & 0.72 & $\times$0.52 & 18.06  & 19.73 & 20.26 & 21.67 & $\times$0.97 \\
        \midrule
        Expert API & 0.87 & 21.73 & 165.0 & 33.13 & $\times$29.08 & 25.42 & 29.02 & 37.89 & 30.27 & $\times$1.49 \\
        \midrule
        Llamdex & 0.07 & 0.80 & 5.27 & 1.45 & $\times$1.00 & 17.96 & 19.29 & 21.82 & 23.29 & $\times$1.00 \\
        \bottomrule
    \end{tabular}
    \begin{tablenotes}
        \item \textsuperscript{1}Mean relative value compared with Llamdex across all datasets.
        \item \textsuperscript{2}This category encompasses LoRA with PATE-GAN, SeqPATE, PromptPATE, Table Diffusion, and DP-OPT.
    \end{tablenotes}
\end{table*}

\begin{figure*}[t!]
    \centering
    \includegraphics[width=0.24\textwidth]{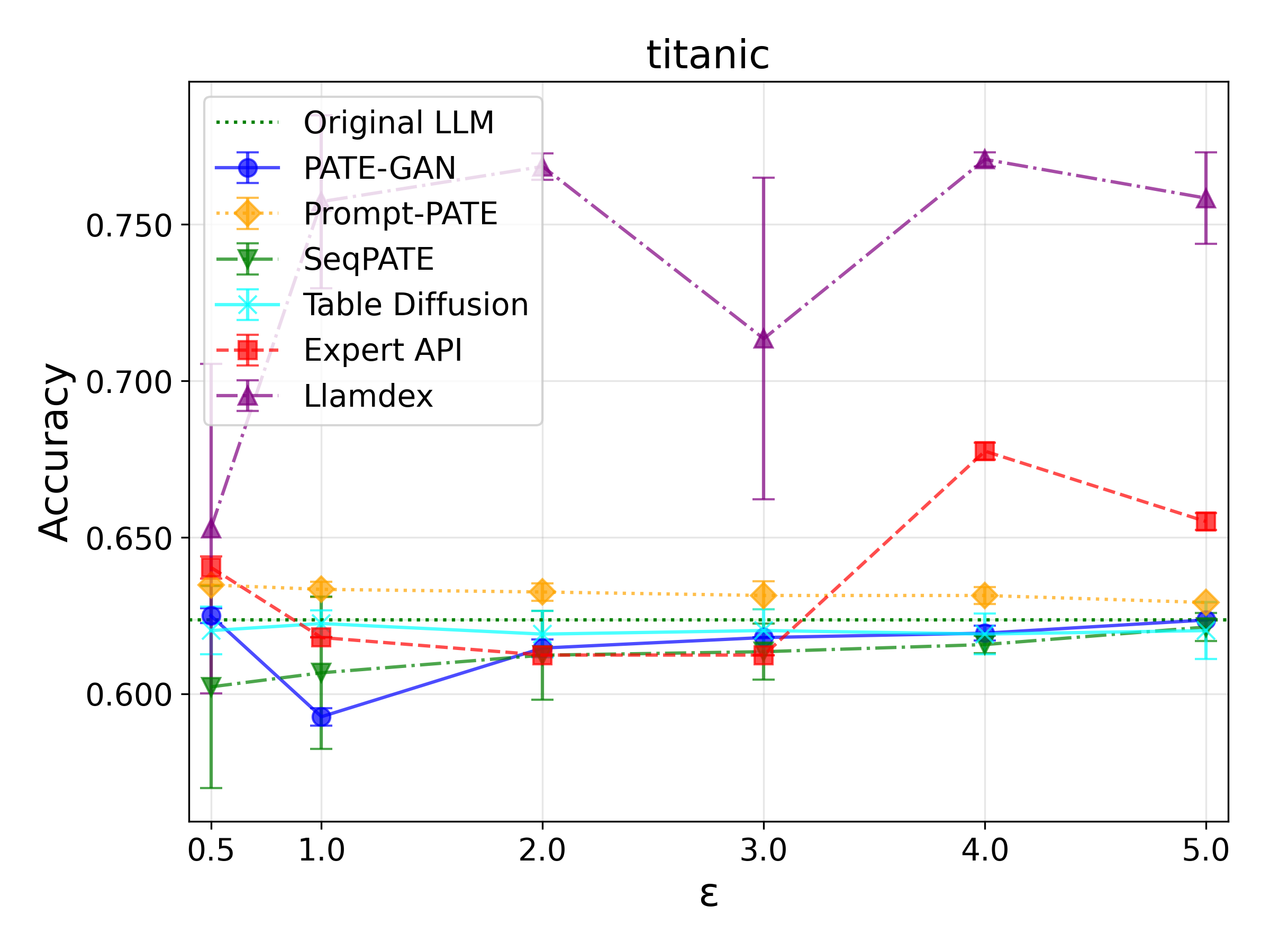}
    \includegraphics[width=0.24\textwidth]{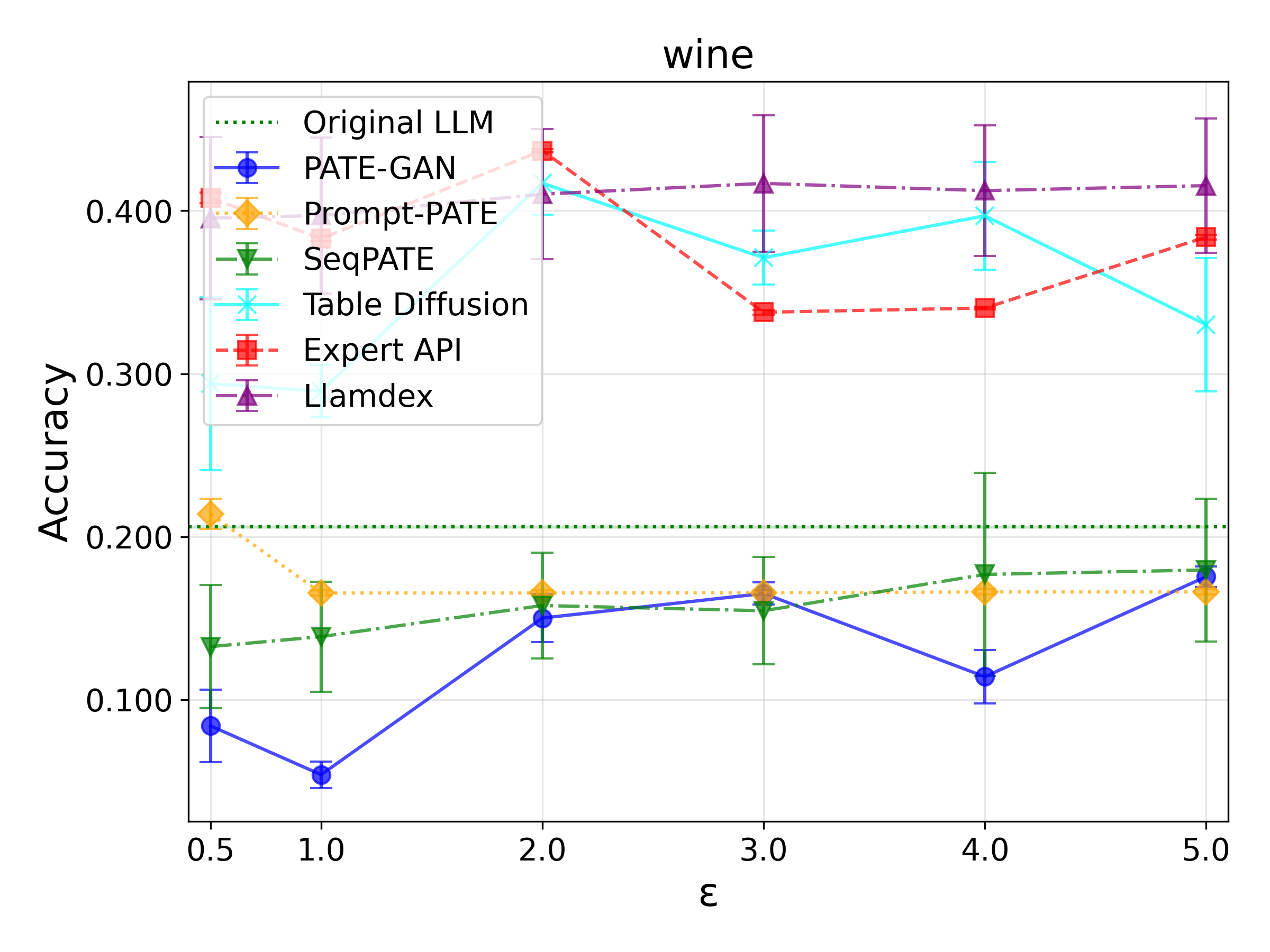}
    \includegraphics[width=0.24\textwidth]{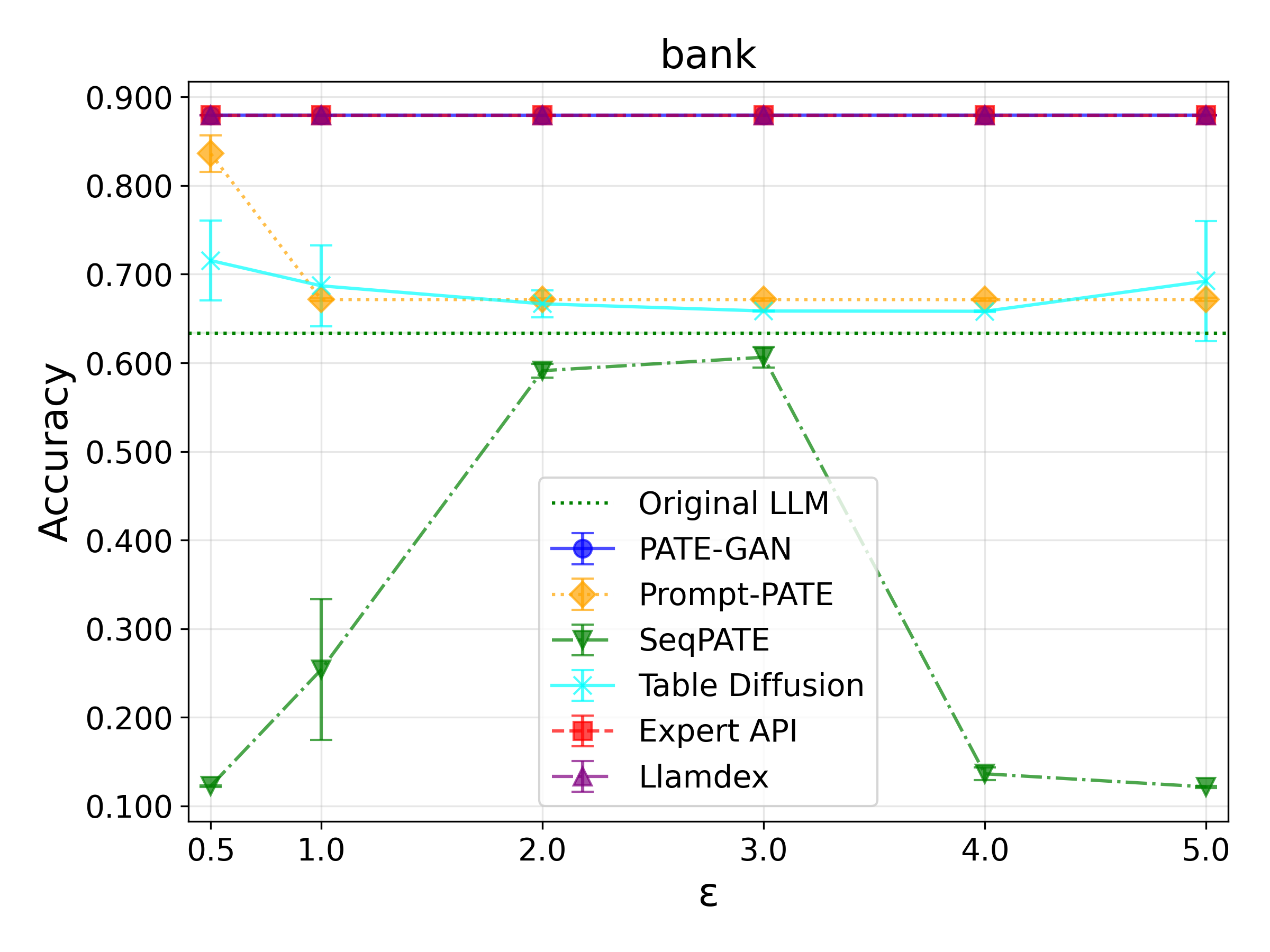}
    \includegraphics[width=0.24\textwidth]{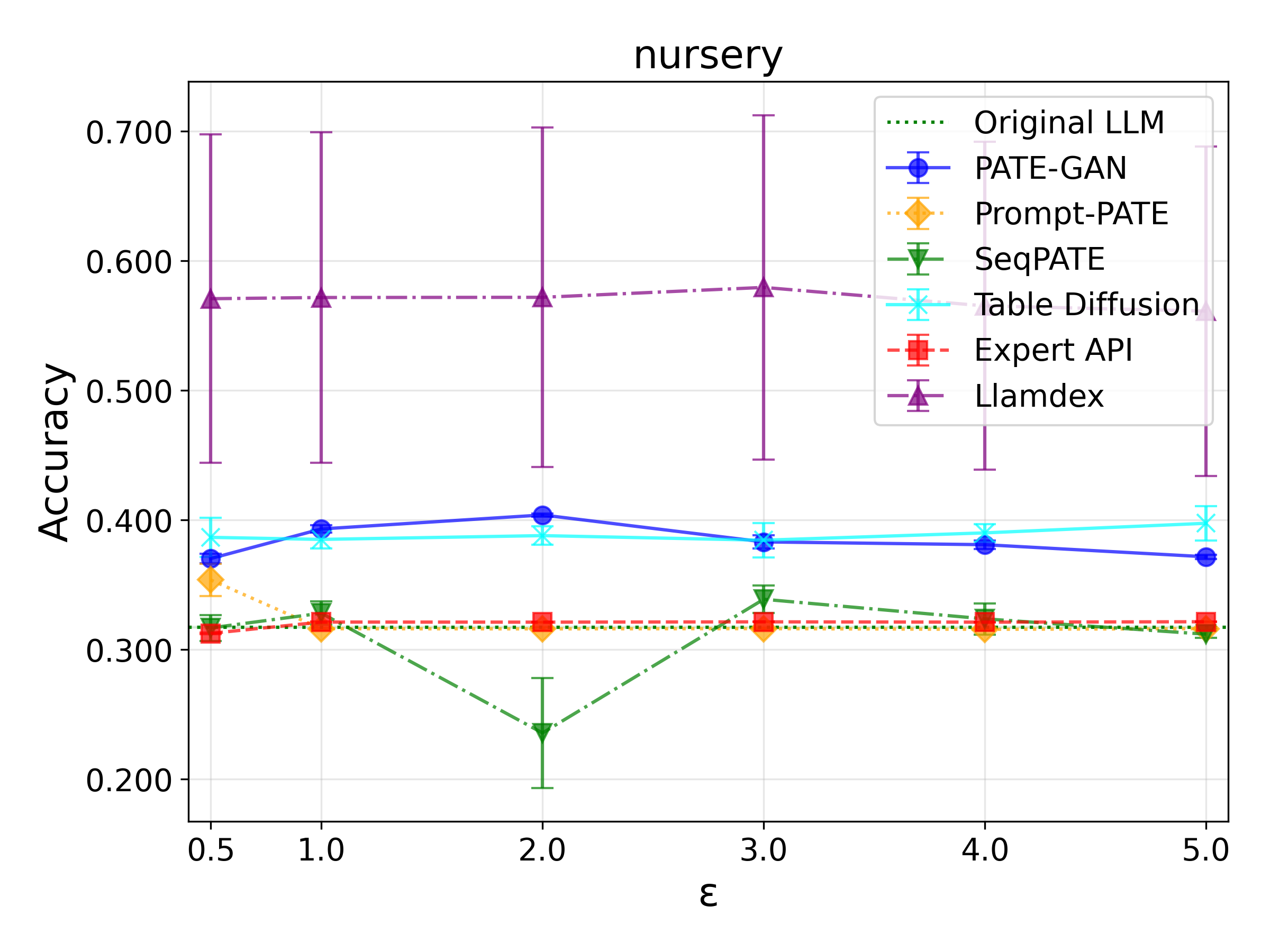}
    \caption{Accuracy of Llamdex and the baselines under different privacy budget $\varepsilon$}
    \label{fig:eps-acc}
\end{figure*}

\subsection{Effectiveness}\label{sec:exp-accuracy}

Evaluation results without differential privacy noise (Table~\ref{tab:performance}) reveal two key findings. First, \textbf{Llamdex with an MLP expert model significantly outperforms all baselines across all datasets.} Notably, on \texttt{nursery}, it surpassed the next best baseline (LoRA fine-tuning on real data) by 14\%, despite no direct access to real domain data. This superiority arises from Llamdex's Llamdex encoder, specifically trained for accurate and comprehensive feature extraction from noisy natural language - a capability often underdeveloped in fine-tuned LLMs. This demonstrates the strong generalization of Llamdex's synthetic data-based training to real-world domain queries. Second, \textbf{Llamdex with an XGBoost expert also shows competitive effectiveness,} generally exceeding other baselines, though typically with slightly lower accuracy than its MLP counterpart. We attribute this to the hard thresholding in tree-based models like XGBoost, which can be more sensitive to minor inaccuracies in extracted input features, potentially leading to incorrect predictions more readily than an MLP's smoother decision boundaries. Overall, these results validate the Llamdex framework's effectiveness in achieving high accuracy for domain-specific question answering without direct access to sensitive domain data.

\begin{table}[ht]
    \centering
    \small
    \caption{Customization Effectiveness without additional privacy noise (\textbf{bold}: best, \underline{underlined}: second best).}\label{tab:performance}
    \vskip 0.15in
    \setlength{\tabcolsep}{4pt}
    \resizebox{\linewidth}{!}{
    \begin{tabular}{cccccc}
        \toprule
        \multirow{2}{*}{\textbf{Method}} & \multicolumn{4}{c}{\textbf{Accuracy (\%)}} \\
        \cmidrule{2-5}
        & \textbf{titanic} & \textbf{wine} & \textbf{bank} & \textbf{nursery} \\
        \midrule
        Real Data LoRA & 62.14 & 31.38 & 81.24 & 37.40 \\
        Original LLM & 62.36 & 20.59 & 63.35 & 31.71 \\
        Expert API & 51.46 & \underline{40.63} & 85.32 & 32.10 \\
        \midrule
        Llamdex-XGBoost & \underline{72.81} & 33.75 & \underline{87.94} & \underline{40.92} \\
        Llamdex-MLP & \textbf{75.51} & \textbf{41.42} & \textbf{87.94} & \textbf{51.69} \\
            \bottomrule
    \end{tabular}}
\end{table}

\subsection{Efficiency}\label{sec:exp-efficiency}

This subsection evaluates the inference efficiency (time and memory consumption) of Llamdex against baselines, with results in Table~\ref{tab:inference} (training efficiency is in Appendix~\ref{sec:train-efficiency}). Two key observations emerge. First, \textbf{Llamdex achieves significantly faster inference than the Expert API}, offering an average 29$\times$ speedup while maintaining inference times similar to lower-performing baselines like LoRA fine-tuned LLMs and the original LLM. Second, Llamdex's memory consumption is similar to these less accurate baselines and 1.49$\times$ lower than the Expert API. This implies Llamdex's superior accuracy-efficiency trade-off over Expert API.

\subsection{Privacy}\label{sec:exp-privacy}

In this subsection, we evaluate the privacy-accuracy trade-off by comparing accuracy under differential privacy with varying $\varepsilon$. The results, shown in Figure~\ref{fig:eps-acc}, reveal two key observations. First, \textbf{Llamdex consistently outperforms baselines even with increased noise} (smaller $\varepsilon$). For instance, on the \texttt{wine} dataset with $\varepsilon=2$, Llamdex outperforms PATE-GAN and PromptPATE by 26\%. Second, we observe that DP data synthesis provides accuracy close to that of the original LLM on most datasets. This is because the synthetic data usually contains too much noise, limiting the useful information available to the LLM and leading to poorer accuracy. In summary, Llamdex strikes a balance between privacy and utility, maintaining high accuracy while ensuring a strong privacy guarantee.

\begin{figure}[htpb]
    \centering
    \includegraphics[width=0.8\columnwidth]{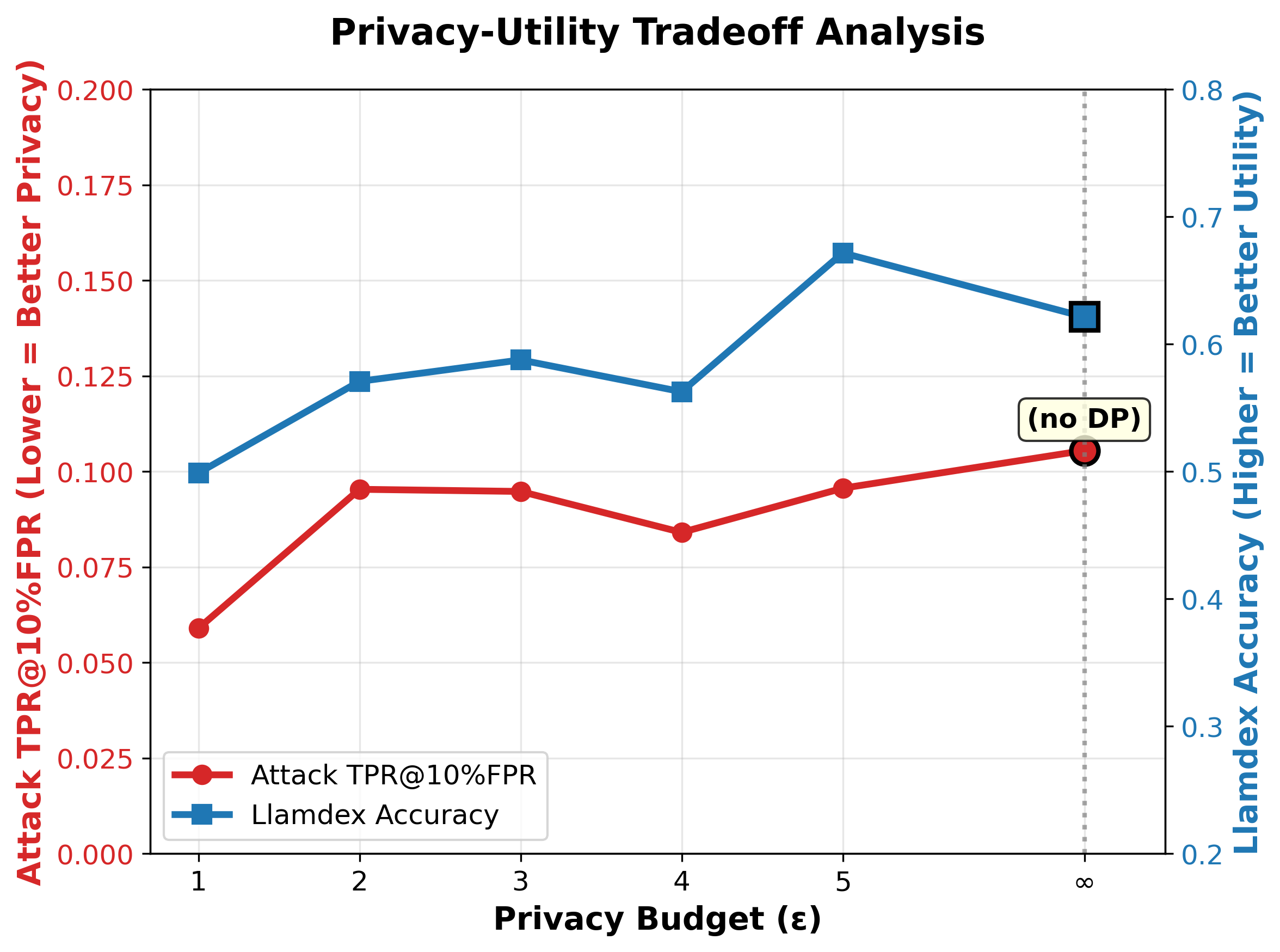}
    \caption{Tradeoff between response accuracy and attack success rate (True Positive Rate at a 10\% False Positive Rate---TPR@10\%FPR)}
    \label{fig:privacy-util-tradeoff}
\end{figure}

\paragraph{Membership Inference Attack.} To assess our model's robustness against advanced membership inference attacks (MIA), we conducted an experiment on the \texttt{titanic} dataset using a state-of-the-art adversary. For a rigorous evaluation, we employed MICO, the winning solution from the official Microsoft Membership Inference Competition, a featured event at the premier IEEE Conference on Secure and Trustworthy Machine Learning (SaTML) 2023. We specifically utilized the champion from the \texttt{Purchase-100} track, designed for tabular data, ensuring a highly relevant and challenging adversary. We evaluated MICO's effectiveness against our expert model trained with varying levels of DP, quantified by the privacy budget $\varepsilon$. The results, presented in Figure~\ref{fig:privacy-util-tradeoff}, are evaluated using attack AUC and the competition's official metric: TPR\@10\% FPR (True Positive Rate at a 10\% False Positive Rate). The findings demonstrate that our DP-protected models are highly effective at mitigating this advanced attack. For instance, at a strong privacy level of $\varepsilon=1.0$, the attack's TPR drops to 0.059, a substantial reduction from the 0.130 achieved against the non-private model. Crucially, this strong privacy protection is achieved while maintaining a high main task accuracy of 0.757. This indicates that our application of DP successfully protects data privacy with minimal impact on model utility, even when tested against a competition-winning attack method.

\subsection{Ablation Study}
This subsection presents ablation studies on the effect of expert weight and the insert layer. Additional ablation studies are detailed in Appendix~\ref{sec:ablation}.

\begin{figure}[htpb]
    \centering
    \includegraphics[width=0.49\linewidth]{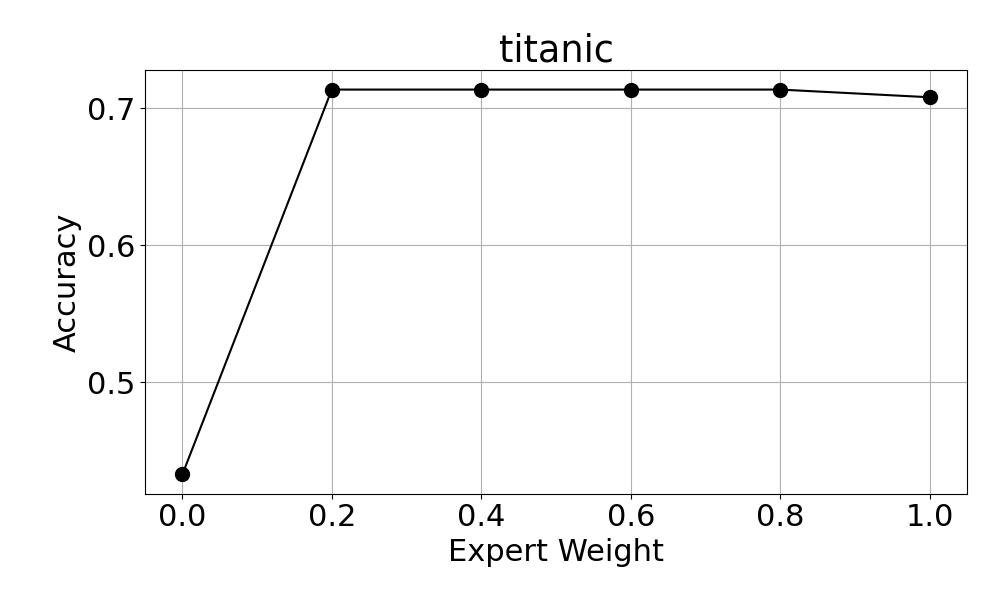}
    \includegraphics[width=0.49\linewidth]{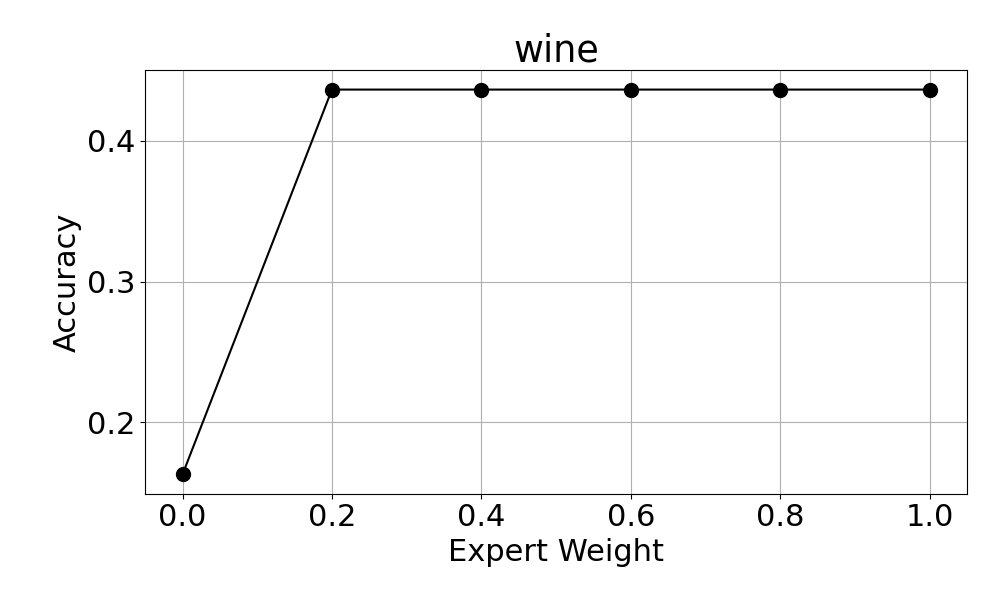}
    \includegraphics[width=0.49\linewidth]{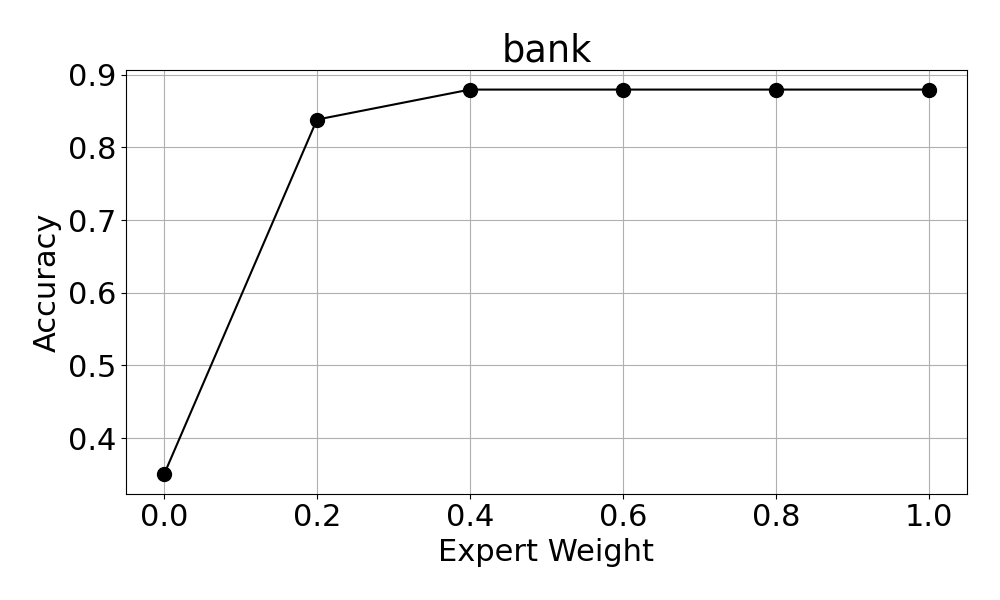}
    \includegraphics[width=0.49\linewidth]{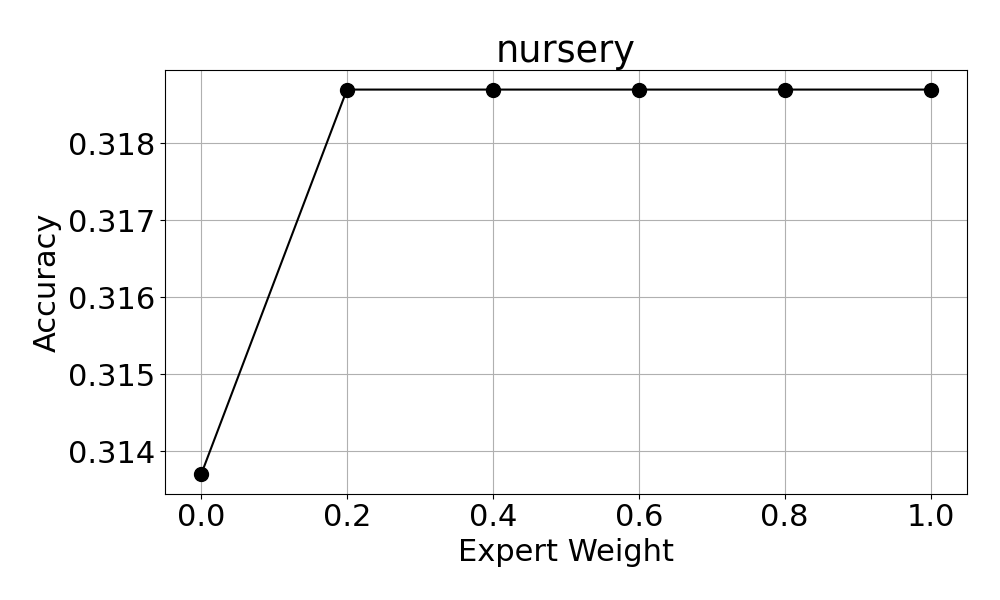}
    
    \caption{Effect of expert weight on Llamdex accuracy}
    \label{fig:llamdex-weight}
\end{figure}

\paragraph{Effect of Expert Weight.}
To demonstrate that the LLM utilizes the expert's output, we scale the expert's output by a weight $\alpha$ and evaluate Llamdex's performance as $\alpha$ varies. The results, shown in Figure~\ref{fig:llamdex-weight}, reveal a positive correlation between Llamdex's performance and the weight $\alpha$, confirming that the LLM leverages the expert's output to enhance prediction accuracy. Notably, when $\alpha=0$, Llamdex's performance drops significantly, underscoring the importance of the expert's output for its effectiveness.

\begin{figure}[htpb]
    \centering
    \includegraphics[width=0.8\columnwidth]{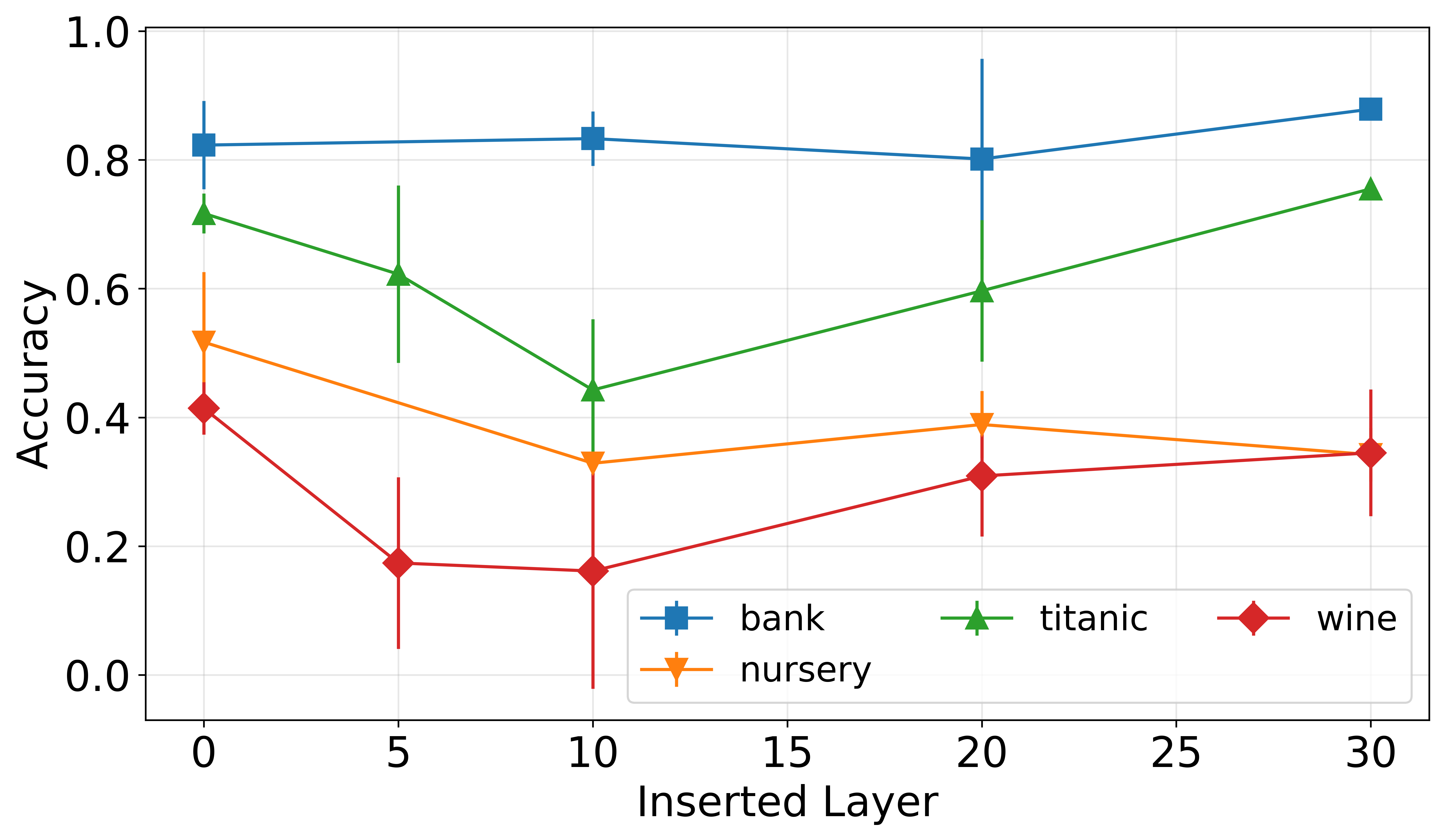}
    \caption{The effect of the depth of the inserted layer on the accuracy of Llamdex}
    \label{fig:llamdex-accuracy-vs-layer}
\end{figure}

\paragraph{Effect of the Insert Layer.}
We explore the accuracy of Llamdex when the expert model is inserted at different layers of the LLM. The effect of the depth of the insertion layer is shown in Figure~\ref{fig:llamdex-accuracy-vs-layer}. From the figure, we observe that the best accuracy occurs when the expert model is inserted in either the first few layers or the last few layers. The probable reason is that the initial and final layers are more closely aligned with natural language tokens, making the information easier to interpret, whereas the intermediate layers are more abstract and harder to map directly to natural language.

\section{Conclusion}\label{sec:conclusion}

We introduced Llamdex, a novel framework for privacy-preserving LLM service customization. Llamdex allows clients to customize LLM service with domain-specific models, thereby preserving the privacy of their private data. Experiments show that Llamdex significantly improves customization effectiveness over baselines without accessing sensitive domain data. By effectively balancing effectiveness, privacy, and inference efficiency, Llamdex provides a robust solution for deploying customized LLM services in sensitive domains.

\section{Limitation}\label{sec:limitation}
While Llamdex demonstrates a novel approach to privacy-preserving LLM service customization, the current work possesses certain limitations that also highlight avenues for future development. Firstly, the framework is primarily designed for interactions involving a single client-provided expert model. Future work could explore mechanisms for dynamic routing and complex architecture to effectively integrate multiple, diverse expert models for multi-task scenarios. Secondly, the present study focuses on customization using tabular data. Extending Llamdex to robustly support multi-modal client data, such as images or unstructured text, which would necessitate distinct Llamdex encoder and decoder architectures, remains a significant area for future research and development.

\section*{Acknowledgement}
This research is supported in part by WeBank Scholars Program Innovative Research 2024. This research is also supported in part by the National Research Foundation, Singapore and Infocomm Media Development Authority under its Trust Tech Funding Initiative. Any opinions, findings and conclusions or recommendations expressed in this material are those of the author(s) and do not reflect the views of National Research Foundation, Singapore and Infocomm Media Development Authority.

\bibliography{references.bib} 

\newpage

\appendix

\part{Appendix} %
\parttoc %
\pagenumbering{gobble}  %

\section{Experimental Details}\label{sec:exp-detail}
In this section, we provide additional details on the datasets used in our experiments and an example of evaluation question.

\paragraph{Dataset Details.} Table~\ref{tab:dataset} presents detailed information (including licenses) about the tabular datasets employed in the experiments, including the additional dataset \texttt{adult} in Appendix~\ref{sec:add-exp}.

\paragraph{Dataset and Library Licensing, and Data Considerations.} The datasets utilized in our experiments are publicly available and are generally provided under permissive licenses. Specifically, the `titanic` dataset is available under a CC0 license. The `wine`, `adult`, `bank`, and `nursery` datasets, commonly sourced from the UCI Machine Learning Repository, are typically distributed under the Creative Commons Attribution 4.0 International (CC BY 4.0) license, which allows for sharing and adaptation with appropriate attribution. For the implementation of differentially private expert model training, we utilize the Opacus library~\cite{yousefpour2021opacus}, which is open-source and licensed under the Apache License 2.0. This permissive licensing for both datasets and key software components facilitates reproducibility and further research. Our work relies on these pre-existing public versions and does not involve re-collection or further direct processing of raw sensitive data containing personally identifying information.

\begin{table*}[htpb]
    \centering
    \caption{Detailed information of tabular datasets, including their licenses.}
    \small %
    \begin{tabular}{lrrrl} %
        \toprule
        \textbf{Dataset} & \textbf{\#Instances} & \textbf{\#Features} & \textbf{\#Classes} & \textbf{License} \\ %
        \midrule
        titanic~\cite{titanic} & 887 & 7 & 2 & CC0 \\ %
        wine~\cite{wine} & 4,898 & 11 & 11 & CC BY 4.0 \\ %
        adult~\cite{adult} & 48,842 & 14 & 2 & CC BY 4.0 \\ %
        bank~\cite{bank} & 45,211 & 16 & 2 & CC BY 4.0 \\ %
        nursery~\cite{nursery} & 12,960 & 8 & 4 & CC BY 4.0 \\ %
        \bottomrule
    \end{tabular}
    \label{tab:dataset} %
\end{table*}

\paragraph{Example of Evaluation.} An example of the tabular data and questions generated from the \texttt{titanic} dataset is shown in Table~\ref{tab:text-data-example}.

\begin{table*}[ht!]
    \renewcommand\tabularxcolumn[1]{m{#1}} 
    \small
    \centering
    \caption{The used prompt and example of tabular data and questions from the \texttt{titanic} dataset in the evaluation}\label{tab:text-data-example}
    \vskip 0.15in
    \begin{tabularx}{\textwidth}{cX}
        \toprule
         & \multicolumn{1}{>{\centering\arraybackslash}X}{\textbf{Example}} \\
        \midrule
        \textbf{Column Names} & Age,  Fare\textsuperscript{1},  Parents/Children Aboard,  Pclass,  Sex,  Siblings/Spouses Aboard,  Survived \\
        \midrule
        \textbf{Tabular Row} & 18.0,\quad 9.35,\quad 1,\quad 3,\quad female,\quad 0,\quad 1 \\
        \midrule
        \textbf{Question Generation Prompt} & Convert the following information about a Titanic passenger into natural language. Ensure and 
        double-check that you do not miss any information, add some irrelevant context, and ask if the passenger survived
        or not at the end without answering, \# please: \#Sex: \textbf{female} \#num\_parents\_and\_children\_aboard: \textbf{1} \#Fare: \textbf{9.35} \#Age: \textbf{18.0} \#num\_siblings\_and\_spouses\_aboard: \textbf{0} \#ticket\_class: \textbf{Third class} \\
        \midrule
        \textbf{Generated Question} & This information pertains to a \textbf{female} passenger aboard the Titanic. She was \textbf{18 years old} and traveled in the \textbf{Third class}. She was accompanied by \textbf{one parent or child}. It is also noteworthy that she did \textbf{not have any siblings or spouses} aboard the ship. Her fare for the journey was \textbf{9.35} dollars. Could you please confirm if this passenger survived the tragic sinking of the Titanic or not? \\
        \midrule
        \textbf{System Prompt} & Respond the user's question in only one word: Yes or No. \\
        \midrule
        \textbf{Answer of Llamdex} & Yes \\
        \bottomrule
    \end{tabularx}
    \begin{tablenotes}
    \item \textsuperscript{1}As the unit of ``fare'' is not defined in the dataset, we interpret it as being in dollars without specifying the currency. Given that \texttt{titanic} serves as an external knowledge base, the actual unit does not affect the evaluation outcome.
    \end{tablenotes}
\end{table*}

\section{Training Efficiency}\label{sec:train-efficiency}
The training time and memory consumption for Llamdex and the baselines are presented in Table~\ref{tab:training}. From the results, we can make one key observation: training Llamdex requires a similar order of magnitude of resources as LoRA fine-tuning, with memory consumption also comparable to LoRA. The slightly higher memory usage and increased training time in Llamdex are due to the additional parameters introduced by the Llamdex encoder and decoder. This indicates a trade-off between efficiency and accuracy, with Llamdex providing a significant improvement in accuracy over LoRA. Expert API and original LLM is not included in the training time comparison as they do not require training.
\begin{table*}[htpb]
    \centering
    \caption{Average Training Time (minutes per epoch) and Peak Training Memory Cost (GB)}
    \label{tab:training}
    \vskip 0.15in
    \resizebox{\textwidth}{!}{
    \begin{tabular}{c cccc cccc}
        \toprule
        \multirow{2}{*}{\textbf{Method}} & \multicolumn{4}{c}{\textbf{Training Time\textsuperscript{1} (min)}} & \multicolumn{4}{c}{\textbf{Training Memory (GB)}} \\
        \cmidrule(lr){2-5} \cmidrule(lr){6-9}
        & \textbf{titanic} & \textbf{wine} & \textbf{bank} & \textbf{nursery} & \textbf{titanic} & \textbf{wine} & \textbf{bank} & \textbf{nursery} \\
        \midrule
        \makecell{Real Data LoRA /\\PATE-GAN LoRA} & 3.94 & 4.76 & 4.28 & 3.98 & 16.05 & 17.89 & 16.90 & 16.07 \\
        \midrule
        Llamdex\textsuperscript{2} & 7.42 / 4.73 & 12.15 / 7.47 & 12.17 / 7.42 & 7.77 / 4.85 & 29.96 & 29.47 & 29.28 & 34.46 \\
        \bottomrule
    \end{tabular}}
    \begin{tablenotes}
        \item \textsuperscript{1}For a fair comparison of efficiency, we fix the number of instances per epoch at 10,000 for all methods.
        \item \textsuperscript{2}The per-epoch training time of Llamdex is reported in the format of (training time of Llamdex encoder)/(training time of Llamdex decoder).
    \end{tablenotes}
\end{table*}

\section{Additional Results}\label{sec:add-exp}
In this section, we present additional experimental results. We first present the results of iterative reasoning, followed by an evaluation of F1 scores. Finally, we compare performance on the \texttt{adult} dataset.

\paragraph{Results of Iterative Reasoning.} We demonstrate the results of iterative reasoning using a specific example. In this example, a connection prompt is used: \texttt{"The expert's answer is \{\quad\}. Regard the expert's answer as a fact. Based on the expert's answer, directly answer the following question:"}. During iterative reasoning, the expert's answer is inserted into the blank and fed back into the original LLM, along with a follow-up prompt provided by the user. The results are shown in Table~\ref{tab:reason}. The observations reveal that Llamdex \textbf{not only accurately predicts the result, consistent with the real data, but also identifies related features such as age and sex.} In contrast, the original Mistral model fails to make an accurate prediction and provides vague answers.

\begin{table*}[t]
    \centering
    \caption{\rev{Example of Llamdex with iterative reasoning on Titanic survival prediction}}\label{tab:reason}
    \vskip 0.1in
    \small
    \resizebox{\textwidth}{!}{
    \begin{tabular}{p{0.2\textwidth}p{0.8\textwidth}}
        \toprule
        \textbf{Question} & This information pertains to a \textbf{female} passenger aboard the Titanic. She was \textbf{18 years old} and traveled in the third class. She was accompanied by one parent or child. It is also noteworthy that she did not have any siblings or spouses aboard the ship. Her fare for the journey was 9.35 dollars. Could you please confirm if this passenger survived the tragic sinking of the Titanic or not? \\
        \midrule
        \textbf{Follow-up Prompt} & Do you know what is the probable result and the reason what factors affect the result? \\
        \midrule
        \textbf{Mistral's Answer} & Based on the information provided, it is not possible to definitively confirm whether this passenger survived the sinking of the Titanic. Third-class passengers were generally less likely to be rescued due to the limited lifeboats available and the priority given to first- and second-class passengers. However, there were some exceptions, and it is always possible that this individual managed to find a place on a lifeboat.

        Factors that affected the survival rate on the Titanic included class, gender, age, and physical ability. Women and children were given priority boarding on lifeboats, and first-class passengers were more likely to be rescued due to their ability to pay for additional seats on lifeboats. Physical ability also played a role, as those who were stronger and more agile were more likely to survive the frigid waters.
        
        It is important to note that the survival rate for third-class passengers was still relatively low, with only about \textbf{20\% surviving} the sinking. However, without more specific information about this passenger, it is impossible to say for certain whether she survived or not. \\
        \midrule
        \textbf{Llamdex's Answer} & Based on the expert's answer that the 18-year-old third-class female passenger \textbf{survived} the Titanic sinking, the probable result is that she was among the survivors. The factors that likely affected the result include:
        
        1. Class: Third-class passengers were less likely to have access to lifeboats, though some secured spots due to \textbf{being women/children} or through other means
        
        2. Age: \textbf{Younger} passengers, especially women and children, had boarding priority
        
        3. Physical condition: Better condition improved survival chances in freezing water
        
        4. Swimming ability: Swimming skills increased survival chances
        
        5. Life jacket: Improved flotation and survival odds
        
        6. Being in a lifeboat: Direct path to survival
        
        7. Being rescued by another ship: Some survivors were picked up by other vessels
        
        8. Proximity to lifeboats at launch: Better chances of boarding
        
        9. Visibility to passing lifeboats: Increased rescue chances
        
        10. Access to floating debris: Improved survival time in water \\
        \midrule
        \textbf{Survived in Dataset?} & \textbf{Yes} \\
        \bottomrule
    \end{tabular}
    }
\end{table*}

\paragraph{Performance under F1 score.} We evaluate Llamdex's performance on additional metrics - F1 score~\cite{rijsbergen1979information} - for binary classification tasks. The results, presented in Table~\ref{tab:performance-metrics}, demonstrate that Llamdex achieves significant improvements in F1 scores on \texttt{titanic} and shows competitive performance on \texttt{bank}. The relatively low F1 scores across all methods on \texttt{bank} are attributed to the dataset's significant class imbalance.

\begin{table}[ht!]
    \centering
    \small
    \caption{F1 scores of different methods}\label{tab:performance-metrics}
    \vskip 0.15in
    \begin{tabular}{ccc}
        \toprule
        \multirow{2}{*}{\textbf{Method}} & \multicolumn{2}{c}{\textbf{F1-Score}} \\
        \cmidrule{2-3}
        & \textbf{bank} & \textbf{titanic} \\
        \midrule
        Real Data LoRA & 0.7988 & 0.5721 \\
        Original LLM & 0.6936 & \underline{0.5738} \\
        Expert API & \textbf{0.8355} & 0.4933 \\
        \midrule
        Llamdex-MLP & \underline{0.8228} & \textbf{0.7578} \\
        \bottomrule
    \end{tabular}
\end{table}

\rev{\paragraph{Performance on \texttt{adult} Dataset.} We evaluate the performance of Llamdex (without DP noise) against the original LLM, Real Data LoRA, and Expert API on the \texttt{adult} dataset, as shown in Table~\ref{tab:performance-adult}. The results indicate that Llamdex outperforms the original LLM and is competitive with the Expert API. The smaller performance improvement of Llamdex on this dataset can be attributed to the simplicity of the schema of \texttt{adult}, which makes it easier for the Expert API to extract accurate values from natural language descriptions.}
\begin{table}[ht!]
    \centering
    \small
    \caption{Accuracy of Llamdex-MLP on adult dataset}\label{tab:performance-adult}
    \vskip 0.15in
    \begin{tabular}{cccc|c}
        \toprule
        \textbf{Dataset} & \makecell{\textbf{Real Data} \\ \textbf{LoRA}} & \makecell{\textbf{Original} \\ \textbf{LLM}} & \makecell{\textbf{Expert} \\ \textbf{API}} & \textbf{Llamdex} \\
        \midrule
        \textbf{adult} & 69.84 & 60.86 & \textbf{76.55} & \underline{76.08} \\
        \bottomrule
    \end{tabular}
\end{table}

\section{Ablation Study}\label{sec:ablation}

\subsection{Effect of Base LLM - Llama-based Llamdex}

We conduct experiments similar to those in Table~\ref{tab:performance} to evaluate performance without additional privacy noise. Llamdex's performance on another base LLM (Llama-2-7B~\cite{touvron2023llama}) is presented in Table~\ref{tab:performance-llama}. Notably, Llamdex with Llama-2-7B demonstrates a significant improvement in accuracy compared to the baselines, indicating that Llamdex is robust to the choice of the base LLM.

\begin{table}[ht!]
    \centering
    \small
    \caption{Accuracy of Llamdex (Llama-2) without additional privacy noise}\label{tab:performance-llama}
    \vskip 0.15in
    \begin{tabular}{ccccc}
        \toprule
        \multirow{2}{*}{\textbf{Method}} & \multicolumn{4}{c}{\textbf{Accuracy (\%)}} \\
        \cmidrule{2-5}
        & \textbf{titanic} & \textbf{wine} & \textbf{bank} & \textbf{nursery} \\
        \midrule
        Real Data LoRA & \underline{59.55} & \underline{1.03} & \underline{85.80} & \textbf{35.09} \\
        Original LLM & 39.33 & 0.00 & 12.14 & 30.98 \\
        \midrule
        Llamdex-MLP & \textbf{75.17} & \textbf{25.91} & \textbf{87.91} & \underline{31.89} \\
        \bottomrule
    \end{tabular}
\end{table}

\subsection{Effect of Number of Tokens}

\begin{figure}
    \centering
    \includegraphics[width=\linewidth]{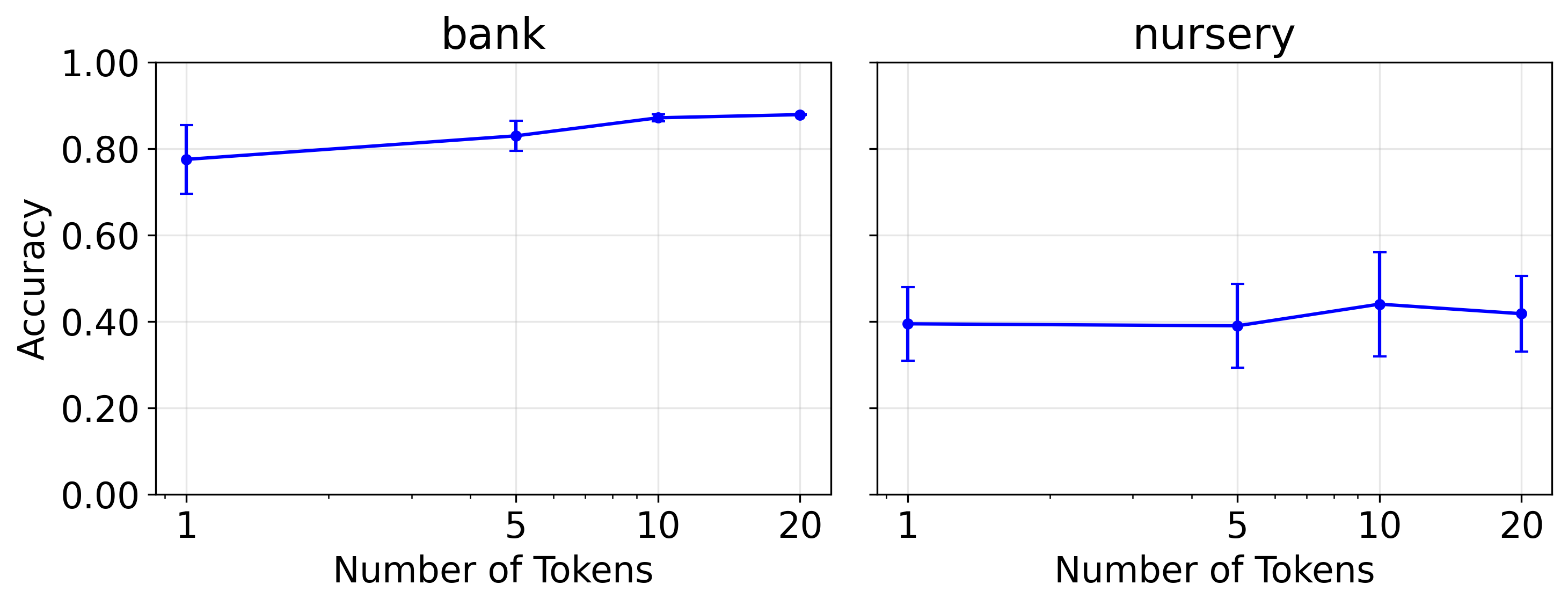}
    \caption{The effect of the number of tokens on the accuracy of Llamdex}
    \label{fig:llamdex-tokens}
\end{figure}

We fix the inserted layer to 0 and vary the number of tokens used to store the expert's output embeddings to evaluate Llamdex's performance. The results, shown in Figure~\ref{fig:llamdex-tokens}, indicate that increasing the number of tokens generally improves accuracy across datasets. 

\subsection{Effect of Token Mapping}

We evaluate the performance of Llamdex with and without token mapping, as shown in Table~\ref{tab:llamdex-token-map}. The results indicate that removing token mapping significantly reduces Llamdex's accuracy, highlighting the importance of the token mapping layer. This is because the token embeddings of one LLM cannot be directly used by another LLM with a different token vocabulary. Token mapping enables the SLM to interpret the LLM's token embeddings, resulting in more accurate predictions.
\begin{table}[ht!]
    \centering
    \caption{Performance Comparison between Llamdex w/ and w/o Token Mapping}
    \label{tab:llamdex-token-map}
    \vskip 0.15in
    \small
    \begin{tabular}{ccc}
        \toprule
        \multirow{2}{*}{\textbf{Dataset}} & \multicolumn{2}{c}{\textbf{Accuracy (\%)}} \\
        \cmidrule{2-3}
        & \textbf{w/o token mapping} & \textbf{w/ token mapping} \\
        \midrule
        titanic & 44.38 & \textbf{75.51} \\
        wine & 12.52 & \textbf{41.42} \\
        bank & 45.57 & \textbf{87.94} \\
        nursery & 32.20 & \textbf{51.69} \\
        \bottomrule
    \end{tabular}
\end{table}

\subsection{Effect of Gaussian Padding.}

We evaluate the performance of Llamdex with zero padding and Gaussian padding, as shown in Table~\ref{tab:llamdex-zero}. The results show that Gaussian padding significantly improves Llamdex's accuracy compared to zero padding. This improvement occurs because Gaussian padding breaks the symmetry of parameters, facilitating more effective learning, similar to the model's parameter initialization.

\begin{table}[ht!]
    \centering\small
    \caption{Performance of Llamdex w/ Zero Padding and Gaussian Padding}
    \label{tab:llamdex-zero}
    \vskip 0.15in
    \begin{tabular}{ccc}
        \toprule
        \multirow{2}{*}{\textbf{Dataset}} & \multicolumn{2}{c}{\textbf{Accuracy (\%)}} \\
        \cmidrule{2-3}
        & \textbf{Zero Padding} & \textbf{Gaussian Padding} \\
        \midrule
        titanic & 64.83 & \textbf{75.51} \\
        wine & 13.62 & \textbf{41.42} \\
        bank & 85.94 & \textbf{87.94} \\
        nursery & 32.23 & \textbf{51.69} \\
        \bottomrule
    \end{tabular}
\end{table}

\subsection{Effect of Incomplete Schemas}

This subsection evaluates the robustness of Llamdex to inaccurate feature ranges in the schema. We conduct an experiment using the \texttt{titanic} dataset, focusing on the common \texttt{age} feature. In this experiment, we vary the declared range for \texttt{age} in the schema while keeping the underlying data unchanged. As demonstrated in Table~\ref{tab:incorrect_age}, the resulting AUC-ROC score exhibits minimal variance, remaining stable even when the declared range is substantially mis-specified. This resilience is by design, as Llamdex utilizes broad, semantic estimates (e.g., an \texttt{age} range of $[0, 100]$) rather than requiring precise statistical ones.

\begin{table}[htpb]
\centering
\small
\caption{AUC-ROC of Llamdex on the \texttt{titanic} dataset under different ranges for the \texttt{age} feature in the schema}
\label{tab:incorrect_age}
\begin{tabular}{lc}
\toprule
\textbf{Range of \texttt{age} in Schema} & \textbf{AUC-ROC} \\
\midrule
$[0, 40]$   & 0.7940 \\
$[0, 100]$  & 0.8103 \\
$[0, 1000]$ & 0.8141 \\
$[0, 5000]$ & 0.8119 \\
\bottomrule
\end{tabular}
\end{table}

\section{Baseline Hyperparams}\label{sec:baseline-hyper}
In this section, we detail the hyperparameter settings used for each baseline in our experiments.

\subsection{PATE-GAN} 

For PATE-GAN~\cite{jordon2018pate}, we follow the standard hyperparameter configurations as provided in the original GitHub implementation\footnote{\url{https://github.com/BorealisAI/private-data-generation}}. For models employing noisy Stochastic Gradient Descent (SGD), this involves setting the Gaussian noise variance multiplier to 2 and the gradient clipping coefficient to 0.1. Additionally, a micro-batch size of 8 is used, with training carried out over 500 epochs and a general batch size of 64. In the case of the PATE-GAN model, the default configuration specifies an inverse Laplace noise scale multiplier of 0.0001 and a batch size of 64. The model architecture includes 10 teacher discriminators. During each generator iteration in the training process, there are 5 iterations for the teachers and 5 iterations for the student.

\subsection{Prompt-PATE} 

For the Prompt-PATE~\cite{duan2024flocks} baseline, we employ a configuration of 50 teacher models, each consisting of Mistral-7B, which is consistent with the base LLM utilized in our primary experiments. The Confident-GNMax algorithm~\cite{papernot2018scalable} utilizes a confidence threshold of 10.0. Differentially private synthetic data generated by these teacher models is subsequently employed to fine-tune the target Mistral-7B LLM. This fine-tuning process is conducted using Low-Rank Adaptation (LoRA)~\cite{hu2022lora} with a rank $r=16$ and $\alpha=32$. Training proceeds for 10 epochs, with a learning rate of $5 \times 10^{-5}$ and a batch size of 32.

\subsection{SeqPATE}

For SeqPATE~\cite{tian2022seqpate}, we replicate the settings from the original paper \cite{tian2022seqpate}. This configuration uses 10 teacher models (Mistral-7B, consistent with the base LLM in our experimental setup) with a top-$k$ value of 200. The teacher supervision loss weight ($\lambda$) is set to 20, following the original paper. Both teacher and student models utilize Mistral-7B. Their training employs LoRA fine-tuning with a rank $r=16$ and $\alpha=32$. Training for these models spans 10 epochs, using a learning rate of $5 \times 10^{-5}$ and a batch size of 4.

\subsection{TableDiffusion}

TableDiffusion~\cite{truda2023generating} is trained for 5 epochs with a batch size of 1024 and a learning rate of 0.005. The number of diffusion steps is set to 3. The data generated by this method is then used to fine-tune the base LLM. The fine-tuning parameters are identical to those employed for Prompt-PATE (LoRA $r=16$, $\alpha=32$, learning rate $5 \times 10^{-5}$, batch size 32, training for 10 epochs).

\subsection{DP-OPT} 

For DP-OPT~\cite{hong2024dpopt}, we adhere to the default parameter settings from its original code repository. This involves generating 5 prompts per data point. The LLM is configured to read a maximum of 5 demonstrations at a time, and generated prompts possess a maximum token length of 128. An ensemble of 205 LLMs (\texttt{ensemble\_num}) is utilized. The generation temperature is set to 0.7, with only one generation round performed. Poisson sampling is employed with $q=0.1$. To mitigate the generation of repetitive, low-quality text, a repetition penalty of 1.2 is applied. 10\% of the training data is reserved for validation. Considering the potential for the LimitedDomain mechanism in DP-OPT to fail, a maximum of 20 failures per prompt (with retries) are permitted before prompt generation is terminated. The local model is Llama-3.1-8B-Instruct~\cite{grattafiori2024llama}, and the server model is Mistral-7B-Instruct-v0.3, aligning with the base LLM configuration in our experiments. The initial prompts for each dataset are as follows:

\begin{itemize}[leftmargin=*] %
    \item \textbf{titanic \& bank\_marketing:} \texttt{Answer the following question. Your answer MUST be either 0 (No) or 1 (Yes). Enclose ONLY the integer in \textbackslash boxed\{...\}.}
    \item \textbf{wine\_quality:} \texttt{Answer the following question. Your answer MUST be an integer between 0 and 10 (inclusive), where a larger integer indicates better wine quality. Enclose ONLY the integer in \textbackslash boxed\{...\}.}
    \item \textbf{nursery:} \texttt{Answer the following question. Your answer MUST be an integer (0, 1, 2, or 3) corresponding to one of these categories: 0 – special priority (e.g. veterans, siblings), 1 – priority (e.g. staff children, local), 2 – very recommended (strong applicants), 3 – not recommended (weak applicants). Enclose ONLY the integer in \textbackslash boxed\{...\}.}
\end{itemize}

During evaluation, the generated output is considered correct if the text within the \texttt{\textbackslash boxed\{\}} environment matches the expected answer (after removing any leading or trailing whitespace). If the \texttt{\textbackslash boxed\{\}} environment is not found, the prediction is treated as a chance probability.

\subsection{Expert API} 

The Expert API baseline evaluates the LLM's capacity to interact with an external expert model by generating structured API calls. The LLM is provided with a detailed prompt that first describes the interface of the expert model. This description includes the precise JSON format specifying the structure of the expert model's inputs (features it accepts) and outputs (the predictions it returns). Subsequently, the prompt instructs the LLM on the specific string format required to construct a query for this expert model, based on the user's question. This query string is expected to be a sequence of feature-value pairs, clearly delineated. The template of the prompt provided to the LLM is as follows:

\begin{flushleft}
\small %
\ttfamily %
You are a data analysis assistant who strictly adheres to instructions. You have access to a model. It receives input in the following json format:\newline
<JSON input/output format for the expert model is detailed here.>
\newline
Now you should generate a query to answer the user's question. Use the following format to generate queries:\newline
<Specific query string format (e.g., "feature1: [value1] feature2: [value2]...") is detailed here.>
\end{flushleft}

The generated text from the LLM, which should contain the formatted API call, is then parsed. This extracted call is used to query the external expert model. A prediction is considered correct if the response from the expert model aligns with the expected answer.

\section{Future Directions}\label{sec:future-work}
In this section, we discuss the potential future extensions of the Llamdex service and the major challenges involved.

\paragraph{Multi-Task Llamdex.} In real-world applications, more complex scenarios may arise where a single user question requires input from multiple client-provided expert models. For instance, in the medical domain, a question about a patient's symptoms might necessitate inferences from various diagnostic models, such as a radiology model, a pathology model, and a clinical model. The primary challenge in extending the Llamdex service to support multiple client-defined tasks lies in token routing. Similar to Mixture of Experts (MoE) in LLMs, a gating module would be required to determine which tokens (representing parts of the user query or intermediate states) should be routed to which client expert model. To support such multi-task customization, beyond the existing Llamdex encoder and decoder design, the Llamdex framework must also incorporate a carefully designed gating module for efficient token routing. This is left for future work.

\paragraph{Complex Questions.} In practice, user questions are often more complex and may require multiple processing steps to arrive at a comprehensive answer. For example, a question might first require inferring a diagnosis from a radiology image (using one client expert model) before using that diagnosis to formulate a response (potentially involving another expert model or the base LLM's reasoning capabilities). While API-based approaches often utilize chain-of-thought reasoning or Depth First Search-based Decision Trees (DFSDT) to handle such complex queries, these methods, as demonstrated in our experiments, face significant efficiency challenges. A potentially more efficient approach for the Llamdex service could involve integrating client expert models at different layers of the base LLM, enabling the customized service to handle complex questions more effectively. We leave this extension for future work.

\paragraph{Multi-Modal Llamdex.} In this paper, our focus is on tabular data, which is commonly found in relational databases. Potentially, the Llamdex service can be extended to support client customization with multi-modal data, such as images and text. Integrating these multi-modal data sources would necessitate different interfaces for the client's expert models, which in turn may require distinct designs for the Llamdex encoder and decoder components. For example, for image data, the Llamdex encoder might incorporate a convolutional neural network to process visual features for the client's image-based expert model, while for text data, it could utilize attention layers tailored to textual expert models. We leave this extension for future work.

\section{Discussion}\label{sec:discussion}

This section discusses the use of AI assistants in this study and the potential risk of Llamdex.

\paragraph{Use of AI Assistants in Research and Development.}
The development of this research and the preparation of this manuscript were aided by AI-powered assistants. Specifically, Google's Gemini was utilized to assist with aspects of writing to ensure clarity in explanations. For coding tasks related to the implementation and experimentation of the Llamdex framework, Cursor, an AI-assisted code editor, was employed to enhance productivity and aid in code generation and debugging.

\paragraph{Potential Risks.}
The optional use of Differential Privacy (DP), particularly DP-SGD, for training the client-side expert model within the Llamdex framework prompts a broader discussion of such privacy-enhancing technologies. It is important to recognize that DP, as a general mechanism, has inherent limitations applicable to any system employing it. DP does not eliminate all privacy risks entirely; rather, it quantifies and reduces them to an acceptable level defined by the privacy budget (typically $\varepsilon$ and $\delta$). The selection of these parameters is critical: a very small $\varepsilon$ provides stronger privacy but often leads to greater noise injection and, consequently, lower model utility. Conversely, a larger $\varepsilon$ may yield better utility but offers weaker privacy guarantees. Therefore, the practical deployment of DP in any context, including its optional use with Llamdex, necessitates a careful and context-dependent calibration of the privacy budget. Implementers must consider the sensitivity of the data, the specific threat model, and the acceptable trade-off between privacy and model performance. Misconfiguration of DP parameters could lead to either inadequate privacy protection or a significantly degraded user experience. Llamdex's modular design for integrating client-provided models is advantageous in this regard, as it can be readily adapted to incorporate alternative or future privacy-enhancing technologies beyond DP as they become available and mature. Future work in the broader field could explore adaptive or automated methods for selecting optimal DP parameters based on data characteristics and specific application requirements to mitigate these challenges.

\end{document}